\documentclass[preprint,12pt]{elsevier_latex/elsarticle}

\usepackage{amssymb}
\usepackage{amsmath}
\usepackage{booktabs}
\usepackage{graphicx}
\usepackage{multicol}
\usepackage[bookmarks=true]{hyperref}
\usepackage{algorithmic}
\usepackage{mathtools}
\usepackage{amsmath}
\usepackage{amsfonts}
\usepackage{textcomp}
\usepackage{booktabs}
\usepackage[dvipsnames]{xcolor}
\usepackage{listings}
\usepackage{hyperref}
\usepackage{todonotes}
\usepackage{flushend}
\usepackage{lipsum}  
\usepackage{subcaption}
\usepackage{blindtext}
\usepackage{placeins}
\usepackage{listings}
\usepackage{todonotes}
\usepackage{hyperref}
\usepackage{amsmath}
\usepackage{float}
\usepackage{soul}
\usepackage{placeins}
\usepackage{caption}
\usepackage{subcaption}
\usepackage{natbib}
\usepackage{hyperref} 

\journal{Robotics and Autonomous Systems}

\begin{document}

\begin{frontmatter}


\title{From Simulation to Field: Learning Terrain Traversability for Real-World Deployment}


\author[inst1]{Fetullah Atas}
\ead{fetulahatas1@gmail.com}

\author[inst2]{Grzegorz Cielniak}
\ead{gcielniak@lincoln.ac.uk}

\author[inst1]{Lars Grimstad}
\ead{lars.grimstad@nmbu.no}

\affiliation[inst1]{organization={Faculty of Science and Technology},
            city={Ås},
            postcode={1433}, 
            country={Norway}}
 
\affiliation[inst2]{organization={Lincoln Centre for Centre for Autonomous Systems},
            addressline={Isaac Newton Building}, 
            city={Lincoln},
            postcode={LN6 7TS}, 
            state={Lincolnshire},
            country={UK}}

\begin{abstract}

The challenge of traversability estimation is a crucial aspect of autonomous navigation in unstructured outdoor environments such as forests. It involves determining whether certain areas are passable or risky for robots, taking into account factors like terrain irregularities, slopes, and potential obstacles. The majority of current methods for traversability estimation operate on the assumption of an offline computation, overlooking the significant influence of the robot's heading direction on accurate traversability estimates. In this work, we introduce a deep neural network that uses detailed geometric environmental data together with the robot's recent movement characteristics. This fusion enables the generation of robot direction awareness and continuous traversability estimates, essential for enhancing robot autonomy in challenging terrains like dense forests. To fulfill the data needs of the proposed deep neural network, we present an innovative pipeline that utilizes high-fidelity simulations for the automated generation and labeling of data, dramatically reducing the need for exhaustive human involvement in data collection and labeling. 

The efficacy and significance of our approach are underscored by experiments conducted on both simulated and real robotic platforms in various environments, yielding quantitatively superior performance results compared to existing methods. Moreover, we demonstrate that our method, trained exclusively in a high-fidelity simulated setting, can accurately predict traversability in real-world applications without any real data collection. Our experiments showcase the advantages of our method for optimizing path-planning and exploration tasks within difficult outdoor environments, underscoring its practicality for effective, real-world robotic navigation. In the spirit of collaborative advancement, we have made the code implementation available to the public. 
\end{abstract}


\begin{highlights}

\item \textbf{A Novel Traversibility Estimation Neural Network}: Our proposed deep neural network is specially designed to process raw LIDAR and IMU data, enabling continuous traversability assessments with spatiotemporal point cloud maps. This approach showcases our network's ability to adapt to complex environments without needing real-world data. 

\item \textbf{Automated Data Generation and Labeling}: We have pioneered a high-fidelity simulation pipeline for autonomous data generation and labeling based on the robot's experiences across different terrains, significantly reducing human intervention and facilitating large-scale data processing for neural network training.

\item \textbf{Real-World Validation}: Experiments on both simulated and real-world platforms demonstrate the method's superior predictive accuracy, validating its efficacy in practical scenarios without prior real-world data collection.

\item \textbf{Application and Open-Sourcing}: The application of our traversability assessments in 3D path planning algorithms confirms the method's reliability in generating collision-free paths. To further research and practical application, we have open-sourced our neural network and mapping code.

\end{highlights}


\end{frontmatter}


\section{Introduction}
\label{sec:Introduction}

Field robotics often confronts challenges when operating within unstructured terrains as robustly determining the safe regions in such diverse fields is a non-trivial problem. Such challenges arise in various applications, spanning mining, exploration, defense, agriculture, search and rescue, and beyond. The essence of traversability analysis is to determine safe regions within the environment robustly. Traversability assessment is a critical element for several key operations of autonomous robots, such as formulation and refinement of navigation plans. This study focuses on the robust traversability assessment for robots in challenging outdoor settings, like forests or lightly structured urban areas.

The task of evaluating traversability is inherently complex due to its dependency on various factors, such as the robot's design, the robot's direction (which continuously changes), and the characteristics of the terrain. Consequently, a data-driven approach to traversability estimation requires careful consideration. For example, accurate estimations for one set of robot specifications may become obsolete if there are alterations in the robot's tire dimensions or wheelbase. Acknowledging these dynamics, the methodology for data-driven traversability must be adaptable rather than fixed to account for potential changes in a robot's capabilities.

Historically, researchers have used sensor modalities ranging from cameras and LIDARs to proprioceptive sensing (e.g., IMU) to discriminate between traversable and non-traversable regions. Each sensing technology presents its own set of strengths and limitations. Our study uses a blend of LIDAR and IMU data to achieve an accurate, continuous value for determining traversability costs rather than a simple binary classification. This blend, combined with an automated data generation and labeling strategy that exploits the robot's locomotion experience over the terrains, enables our proposed neural network to generalize well to output traversability estimates that are aware of the robot's angle of approach, which we term as directionality-aware.

Wallin et al.~\cite{Wallin2022} recently proposed a traversability estimation technique using simulated interactions between robots and terrain to produce training data and labels autonomously. Their method employs a neural network trained on a Digital Elevation Map (DEM), enabling robots to navigate a simulated environment, collect data, and refine their learning—a process verified in real-world conditions.

Building upon this foundation, our research advances the state-of-the-art by introducing a data-efficient neural network that directly processes dense point cloud maps and IMU data. The advantage of using point clouds lies in their ability to accurately represent any real-world structure, unlike the DEM that the previous method~\cite{Wallin2022} depended on. This enhancement paves the way for more precise and versatile traversability assessments in robotic applications, which we demonstrate in the \autoref{sec:Experiments}
 
Our approach enhances the concept of automated data generation and labeling by utilizing raw sensory data from LIDAR and IMU, facilitating a seamless transition from simulated environments to real-world terrain. This direct sensor data utilization addresses the constraints of DEM-based methods, which can be limited in overlapping surfaces. Due to their single-height limitation per location, DEMs often overlook potentially navigable spaces beneath tree canopies in scenarios like dense forests. Our method, by contrast, presents a more comprehensive solution for autonomous navigation by accounting for the full spectrum of environmental features that may otherwise be concealed or inaccurately represented in DEMs.

The concern regarding LIDAR's sparse output (compared to an RGB image) is a significant one, as it traditionally hinders the ability to generate granular traversability assessments. Addressing this issue, our technique synthesizes a densified, precise local point cloud map through the integration of an Extended Kalman Filter (EKF) state estimate and a GPU-powered Iterative Closest Point (ICP) algorithm~\cite{Koide2021, Vizzo2023}. This enriched, robot-centric map is critical in our methodology's training and inference phases, ensuring that our system can make fine-grained, accurate traversability estimations beyond mere binary classifications.

Our research introduces the following core contributions in the realm of traversability estimation for autonomous robot navigation through challenging environments:

\begin{enumerate}

    \item We propose a novel deep neural network architecture tailored for traversability estimation based on point cloud data alongside IMU information. This architecture accommodates various point cloud attributes such as normals and curvature, enhancing its adaptability to capture different terrain characteristics.

   \item We have established an automatic data generation and a "proxy" data labeling pipeline that leverages high-fidelity simulations and the robot's terrain locomotion experience, significantly reducing the necessity of human labor in data collection and labeling.

  \item We carry out thorough experiments utilizing both synthetic simulations and real-world data, demonstrating the effectiveness of our approach in real-world scenarios, particularly for path-planning tasks.

 \item Unlike most existing methods, we supply the implementation of our method. \footnote{\href{https://github.com/jediofgever/vox_nav_slam}{https://github.com/jediofgever/vox\_nav\_slam}}. \footnote{\href{https://github.com/jediofgever/traversablity_estimation_net}{https://github.com/jediofgever/traversablity\_estimation\_net}}. 

\end{enumerate}

\section{Related Work}
\label{sec:realted_work}

This section provides an overview of seminal contributions on traversability estimation, categorized according to their methodological taxonomy.

\subsection{Vision-Based Approaches}

Initial work on traversability assessment primarily depended on vision-based methodologies using traditional techniques. Stereo vision was foundational for these methods to estimate the scene's depth.

The authors in~\cite{Manduchi2005} introduced a classification algorithm leveraging stereo-range data. This method considered both surface reflectivity and geometric features derived from range measurements. The authors pointed out the necessity for more comprehensive data to address variations in illumination. Bajracharya et al.~\cite{Bajracharya2013} employed a stereo system to generate dense voxel maps. Their classification approach centered on geometric statistics of the RGBD sensing. They utilized specialized illumination during the robot's nighttime operation to counteract the challenges of low-light conditions.

Prior to the advent of deep learning, traditional machine learning paradigms were at the forefront of research for traversability estimation. Notable contributions include the work of Angelova et al.~\cite{Angelova2007}, who introduced a decision tree classifier that was informed by features extracted from RGB images. This model utilized Speeded-Up Robust Features (SURF) descriptors and incorporated an analysis of various color variations to enhance its prediction performance.
In a related development, Bajracharya et al.~\cite{Bajracharya2008} presented a hybrid approach that combined the strengths of Support Vector Machines (SVM) and naive Bayesian classifiers. This combination sought to leverage the robustness of SVMs and the probabilistic insights offered by naive Bayesian methods, thereby enhancing the traversability classifier's performance in complex environments.

Traditional classifiers predominantly depended on hand-crafted features and descriptors. This reliance posed challenges in terms of adaptability. However, their minimal data requirement was a silver lining compared to deep learning approaches. Kelly~\cite{Kelly2017} compared a feature-driven SVM classifier against a Convolutional Neural Network (CNN) for terrain classification. Their results underscored the superior efficacy of the CNN-based approach.

Recent research has increasingly embraced the advantages of deep learning. For instance, Leung et al. ~\cite{Leung2022} employed a deep CNN for the semantic segmentation of images, enhancing the analysis of terrain traversability. They also utilized data from Digital Elevation Maps (DEM) to identify traversable regions. In a related endeavor, Hosseinpoor et al.~\cite{Hosseinpoor2021} developed a deep semantic segmentation model that processed aerial RGB images emphasizing varying robot modalities. However, the authors did not conduct experiments using either simulated or actual robots.

Bekthi et al.~\cite{Bekhti2020} presented a 2D imagery technique for terrain uniformity evaluation utilized for end-stage traversability assessment. Their methodology integrated motion indicators from inertial sensors with textures captured from RGB images. The combined features derived from both sources were input into a Gaussian process predictor. However, the authors highlighted the method's constraints in vast non-uniform environments, primarily due to significant acceleration readings from the inertial sensor.

Chavez-Garcia et al.~\cite{ChavezGarcia2017} explored terrain traversability through a heightmap-based classification approach. They sourced their training data from terrains that were procedurally generated using Perlin noise. Although the method presented several strengths, it predominantly categorized terrains as simply traversable or non-traversable. This constraint rendered the approach less proficient in recognizing terrains with specific nuanced features, like ramps that are steep yet passable.

\subsection{LIDAR-Based Approaches}

LIDARs, with their precise range measurements, have become a prevalent tool in terrain traversability assessment. In the subsequent sections, we provide a concise overview of recent methodologies that primarily employ LIDARs for estimating traversability.

Agishev et al.~\cite{Agishev2023} introduced a point cloud geometry-based traversability assessment technique to optimize navigation trajectories across unstructured terrains. Their method uses statistical analysis of neighboring points to evaluate the slope and roughness of locations. Despite its strengths, the method lacks directionality awareness, a critical aspect for dynamic environment interaction. Nevertheless, this research is significant as it provides one of the few publicly accessible traversability estimation results and datasets. It offers a valuable baseline for comparative analysis, which we perform in \autoref{sec:Experiments}.

Waibel et al.~\cite{Waibel2022} proposed a traversability estimation methodology employing LIDAR and inertial sensing. They applied statistical methods to gauge traversability measures, reminiscent of our previous work~\cite{Atas2022} and \cite{Agishev2023}, where geometric attributes of local terrain patches were used in assessing the slope and roughness of the terrain.

Martinez et al.~\cite{Martínez2020} showcased a traversablity classification technique for LIDAR data using the Scikit-learn library. The authors highlighted the Random Forest classifier's efficacy in their approach. However, they also pointed out computational inefficiencies in calculating point cloud features.

Thakker et al.~\cite{Thakker2021} put forth a novel method for traversability analysis. Their approach involves positioning the robot in specific poses and then computing metrics derived from the surface point cloud and interactions with the robot's geometric structure. Notably, their method also predicts the robot's stability in these poses. One recognized limitation was the potential misclassification of minor obstacles.

Xue et al.~\cite{Xue2023} unveiled an approach similar to ours in the sense that they also use a robot-centric dense point clouds map. Their method integrates consecutive LIDAR scans to construct a dense temporal map, which they term "LIDAR-based terrain modeling". The traversability is inferred through the geometric relationships between adjacent terrain sections, and they utilize Normal Distribution Transform (NDT) for map generation. Contrasting this with our method, we incorporate inertial sensing and harness dense spatio-temporal point cloud data processed by deep neural networks.
 
\subsection{Hybrid Approaches}

Several approaches integrate RGB images and LIDARs, occasionally further augmented with inertial or tactile sensing. Below, we spotlight some notable methods using multi-modalities.

Wallin et al.~\cite{Wallin2022} introduced a method that can be considered a precursor to our current work. The authors utilized high-resolution topography data for traversability prediction, an approach akin to ours where traversability is modeled as a regression problem. Additionally, they employed simulators for data collection to train their deep neural networks. However, a limitation of this approach is its reliance on a pre-existing map of the environment in real-world scenarios, potentially constraining its practical applicability.

Fusaro et al.~\cite{Fusaro2023} adopted an SVM classifier to gauge the traversability of terrains, integrating both LIDAR and RGB image data to achieve an accuracy of 89.2\%. While this method marks a significant step in terrain analysis, it's important to note that their evaluation was primarily focused on urban contexts, characterized by relatively uniform terrain and infrequent significant elevation changes. This urban-centric approach leaves room for further exploration in more varied and challenging environments.

Building upon the need for more comprehensive terrain analysis, Zhang et al.~\cite{Zhang2019} introduced an approach that extends beyond urban landscapes. Their methodology leverages several point cloud patch features, including roll, pitch, and roughness, combining geometric features while also incorporating the robot's suspension model. This method is particularly notable for its emphasis on enhancing pre-existing motion plans. However, it primarily improves upon existing navigational strategies rather than offering a deepened understanding of terrain traversability per se.

To address the limitations of previous methods in dealing with complex terrains, Haddeler et al.~\cite{Haddeler2022} recently proposed a terrain-probing method that is especially effective in challenging environments like soft soil, puddles, and bushy areas. Their innovative approach bridges geometric and appearance-based sensing, offering a more holistic view of terrain analysis. Despite its advancements, the method has limitations, particularly its reliance on tactile sensing mechanisms, which may restrict its applicability in broader outdoor scenarios.

Guan et al.~\cite{Guan2021} introduced a method for traversability mapping utilizing LIDAR point clouds combined with RGB images. This technique integrated 3D volumetric traversability metrics to develop a global map, primarily to aid in the planning and navigation of an excavator. However, the authors reported limited system testing due to safety concerns. It remains unclear whether the method is apt for long-range navigation, as the paper does not provide explicit details in this regard.

Readers are encouraged to consult the comprehensive survey paper by Borges et al.~\cite{Borges2022}. This work provides an in-depth examination of various traversability analysis methods, categorizing them based on their sensing modalities and taxonomies.

Recent advancements in traversability estimation research increasingly favor deep learning over traditional learning methods. While often requiring less data, traditional techniques face challenges in terrains lacking distinct features, as Kelly~\cite{Kelly2017} points out. In response, our study introduces a deep neural network architecture designed to determine continuous traversability values across varied terrains. We utilize an automated data generation method to address deep learning's data inefficiency based on high-fidelity simulation environments. Additionally, our approach uniquely integrates inertial sensing with spatiotemporal LIDAR point clouds, generating a detailed point cloud map to infer traversability estimates around the robot. These traversability estimation maps enhance local motion planning and control, ensuring safer navigation.

\section{Approach}
\label{sec:Approach}

In this section, we provide an in-depth explanation of our methodology. Our approach harnesses data from LIDAR and IMU sensors to generate detailed, robot-centric traversability point cloud maps.

\begin{figure}[!htb]
    \centering
    \begin{minipage}{1\textwidth}
        \centering
        \includegraphics[width=1\linewidth]{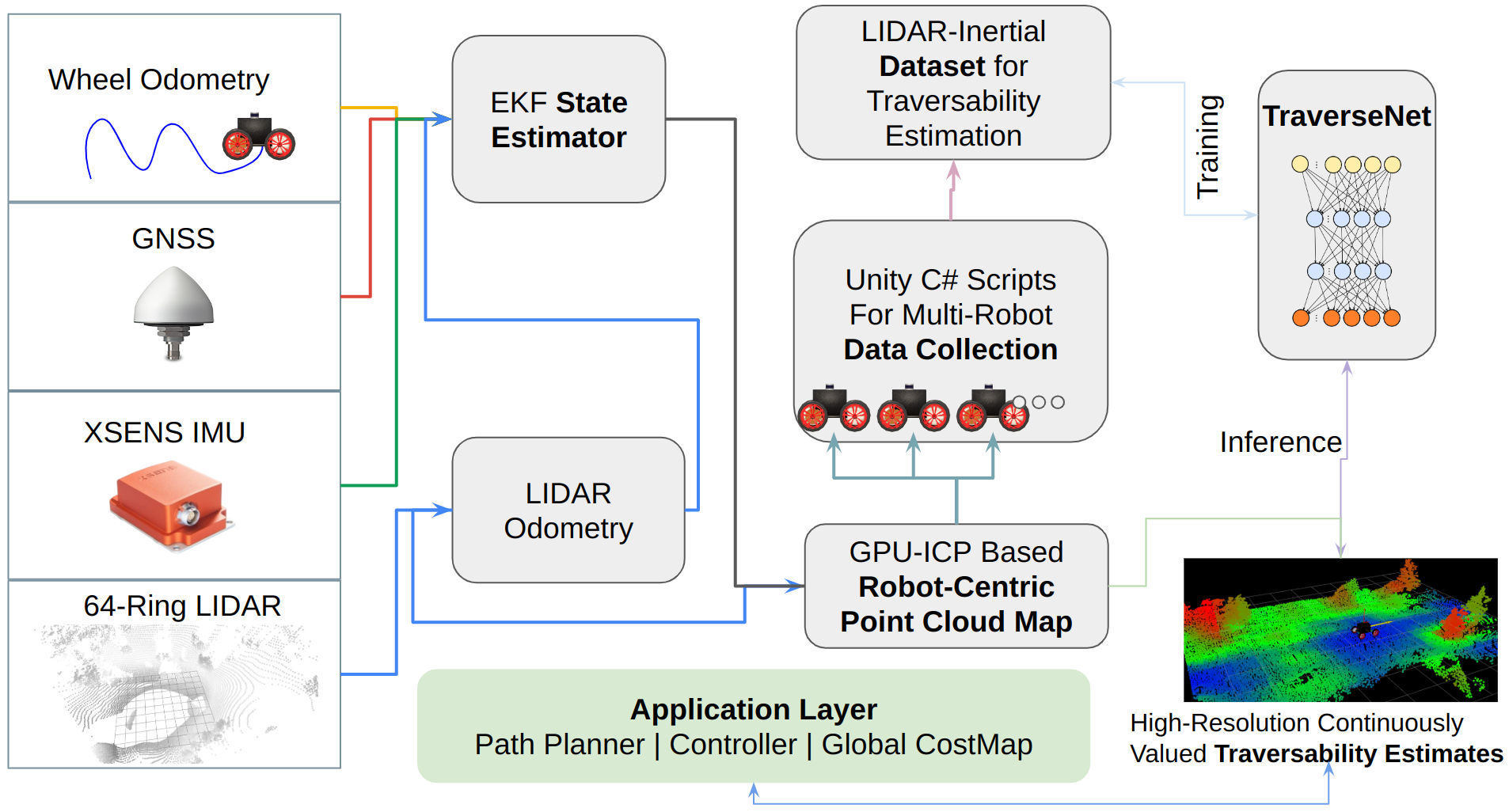} 
        \label{fig:figure1}
    \end{minipage}\hfill
    \caption{Comprehensive Traversability Estimation Pipeline: Utilizing sensor inputs and high-fidelity simulations, data is processed via the EKF State Estimator and fed into TraverseNet. This deep neural network, trained on autonomously generated datasets, outputs precise traversability estimates vital for outdoor autonomous navigation components such as planning, control, and costmap.}
    \label{fig:arhitecture}
\end{figure}

A framework of the proposed method is depicted in \autoref{fig:arhitecture}.
Our system encompasses several key elements: the TraverseNet, a sophisticated LIDAR-Inertial neural network architecture, the auto-generation of a dataset using high-fidelity simulations, and the construction of robot-centric point cloud maps. The following sections will elaborate on each component of our methodology, coupled with a precise problem statement to ensure clarity.

\subsection{Problem Statement}

Our work addresses the challenge of determining traversability for autonomous robots in unstructured environments. We define traversability $T$ as a function that assigns continuous values to the terrain, given the robot's configuration $C$, and local terrain patch state $S$. Precisely, $T$ maps $S$ and $C$ to a continuous value in the range of $[0,1]$:  $T(S, C) \xrightarrow{} [0, 1]$

In this context, $S$ is the local terrain patch state represented by point clouds, which includes spatial properties that affect navigation, and $C$ is the robot's configuration, incorporating attributes such as its orientation and recent locomotion characteristics.

The traversability (cost) score ranges from 0, signifying fully passable, to 1, indicating non-traversable. The task is to estimate $T$ accurately for any pair of $S$ and $C$, necessary for effective path planning and navigation across diverse applications. To tackle this, our approach centers on a deep neural network and an automated data generation system that simulates various terrain states $S$ and robot configurations $C$. This dataset trains our newly proposed neural network $f_\theta$, which aims to approximate the traversability function $T$:
$f_\theta(S, C) \approx T(S, C)$
The network's parameters, $\theta$, are optimized to minimize the discrepancy between the network's output and the actual traversability scores derived from simulations. This optimization enhances the robot's ability to interpret its environment and make informed navigation decisions respecting the terrain traversability.

\subsection{LIDAR-Inertial Traversability Estimation: TraverseNet}

\begin{figure}[!htb]
    \centering
    \begin{minipage}{1\textwidth}
        \centering
        \includegraphics[width=1\linewidth]{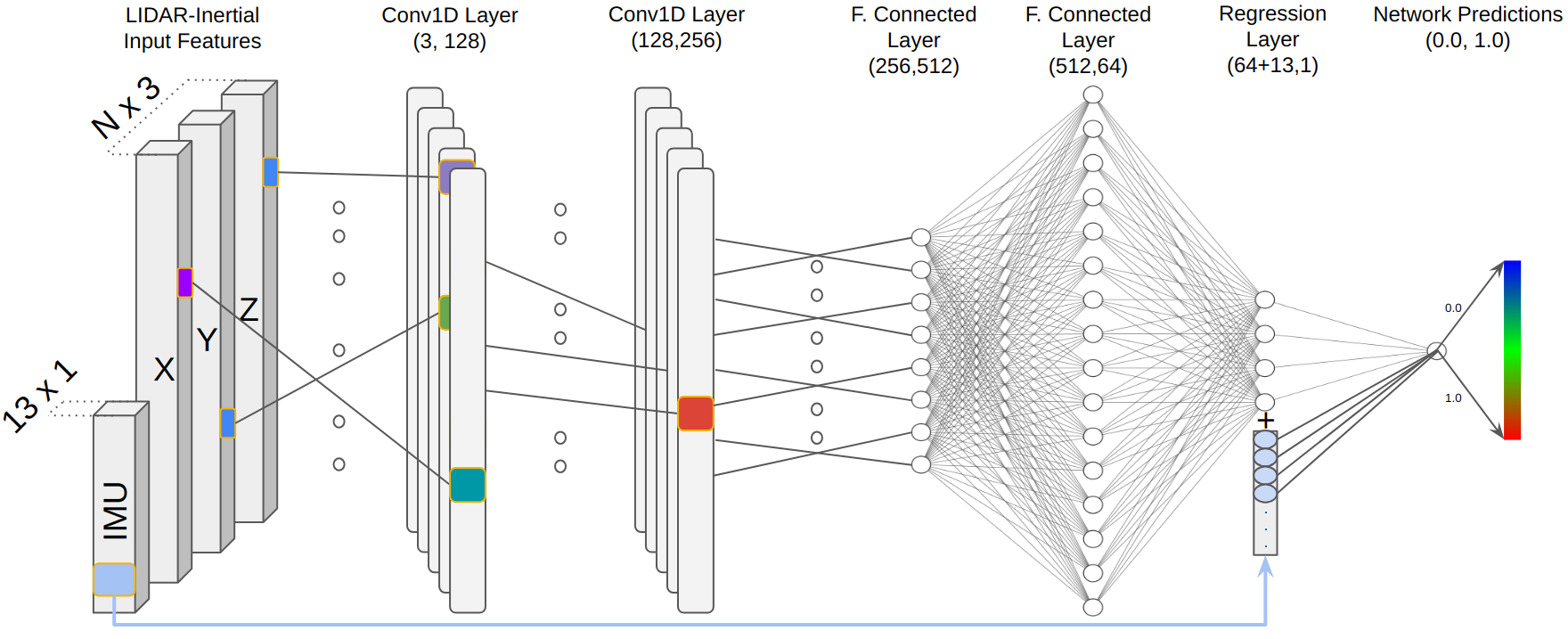} 
        \label{fig:figure1}
    \end{minipage}\hfill
    \caption{Fundamental TraverseNet Design: This illustration presents the core architecture, utilizing point coordinates data for feature extraction through convolutional and fully connected layers, subsequently integrated with a 13-dimensional IMU feature vector. }
    \label{fig:base_traversenet}
\end{figure}

PointNet, introduced by Qi et al.~\cite{PointNet2016} in 2017, revolutionized the handling of raw point clouds by deep learning networks, later leading to a wave of improvements such as PointRCNN~\cite{PointRCNN2019} and PointNet++~\cite{PointNetplusplus2017}. Currently, no PointNet derivatives have been tailored for traversability estimation, an application with promising advantages; Kelly at al.~\cite{Kelly2017} have already reported the superior performance of deep neural nets over the traditional approaches, given that they are provided enough data. Taking a cue from PointNet, our model illustrated in \autoref{fig:base_traversenet} brings a critical enhancement: it is intrinsically sensitive to the robot's \textit{directional orientation}, an aspect frequently overlooked by conventional models. 

For the proposed neural network, we explored various architectural designs, layer sizes, and input feature integration, including direct and intermediary fusion of IMU data. The results, detailed in the accompanying comparison table (\autoref{tab:training_results}), reveal the efficacy of our approach. Our neural network's standard inputs are point features (e.g., coordinates, normal vectors, curvature) alongside a 13-dimensional IMU covariance and orientation vector. We experimented with two main strategies for integrating IMU data: 'direct fusion,' where IMU features are concatenated with latent point cloud features without additional processing (see \autoref{fig:base_traversenet}), and 'mid-fusion,' where IMU features are first processed through convolutional and pooling layers before being combined with the already processed point features (see \autoref{fig:spec_base_traversenet}). Although this paper only details two of such architectures for brevity (see \autoref{fig:base_traversenet} and \autoref{fig:spec_base_traversenet}), comprehensive performance data for various configurations can be found in the experimental results section, see \autoref{tab:training_results}. 

Traversability-aware navigation is a complex task that goes beyond mere obstacle detection; it requires robots to assess how their \textit{directional} movements might affect their ability to traverse different terrains. Traditional geometry-based methods for terrain analysis~\cite{Atas2022,Xue2023}, despite their sophistication, typically do not account for the differences in effort required to move uphill versus downhill, often simplifying traversability into a static measure. This approach neglects the reality that climbing usually requires more power.

Our model advances beyond this limitation by incorporating the robot's angle of approach into traversability calculations, enabling a dynamic evaluation that aligns with the practical challenges encountered across diverse landscapes. This enhancement is achieved through the data labeling technique rooted in the robot's extensive locomotion experience, reflecting the varied demands of different terrain types. The subsequent subsection delves into the details of our locomotion-based data labeling method, shedding light on the way a robot's interaction with its surroundings shapes the nuanced traversability estimates our model provides.

\begin{figure}[H]
    \centering
    \begin{minipage}{1\textwidth}
        \centering
        \includegraphics[width=1\linewidth]{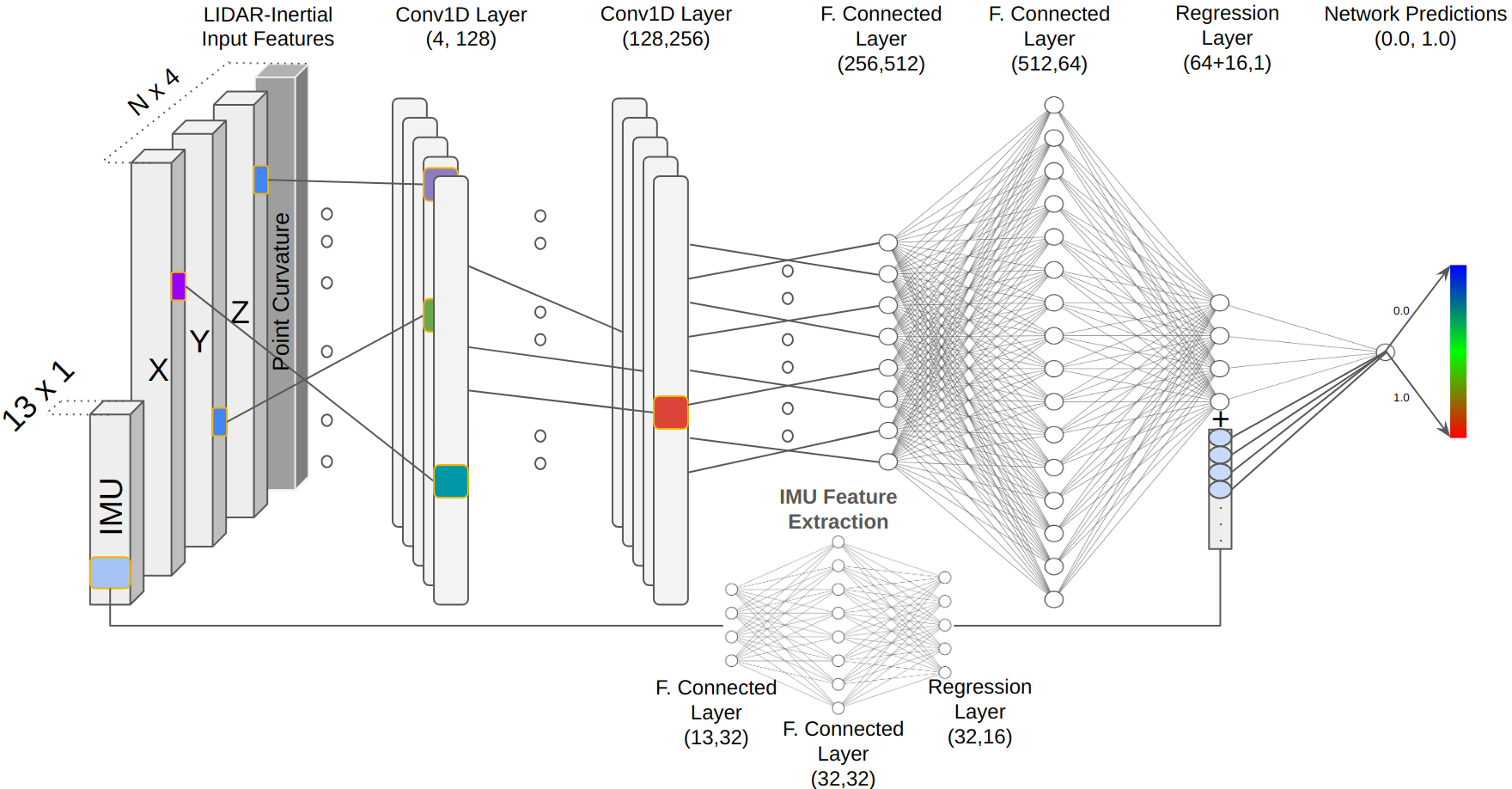} 
        \label{fig:spec_base_traversenet}
    \end{minipage}\hfill
    \caption{Optimal TraverseNet Configuration: Detailed in Section \autoref{sec:Experiments}, the architecture utilizing point coordinates (x, y, z) and point curvature inputs exhibits notable performance. Unlike design with direct IMU fusion depicted in \autoref{fig:base_traversenet}, this approach processes IMU data through additional feature extraction before concatenating it with LIDAR features in the final layer. }
    \label{fig:spec_base_traversenet}
\end{figure}

\subsection{Locomotion as Traversability Indicator}

\begin{figure}[!htb]
    \centering
    \begin{minipage}{1\textwidth}
        \centering
        
        \includegraphics[width=1\linewidth]{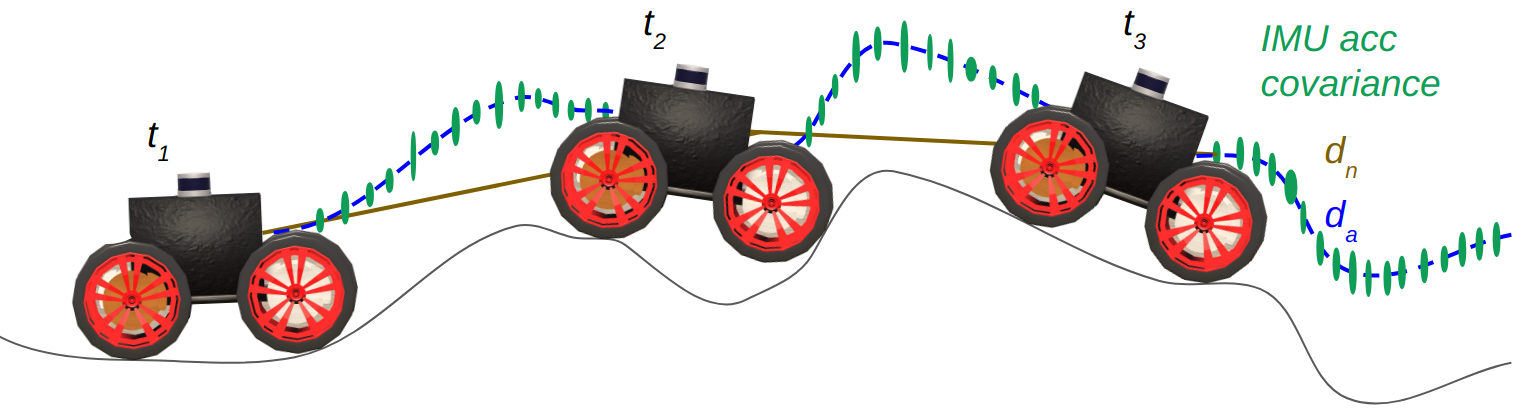} 
        \label{fig:figure1}
    \end{minipage}\hfill
    \caption{An illustration depicting the robot's computation of traversability labels using both nominal and actual travel distances. The blue represents the robot's actual path, the gold symbolizes the anticipated nominal path, and the green displays a time-series representation of the computed IMU acceleration covariance.}
    \label{fig:locomotion}
\end{figure}

Deep learning methods, while effective, necessitate the provision of data and corresponding labels in the context of supervised learning. In our methodology, we treat terrain traversability as a regression problem. Therefore, for each data sample, the neural network requires both the sample and its associated traversability cost label. We utilize the robot's experience across the terrain to fulfill this demand.

With specific physical attributes such as wheel radius and wheelbase, and a defined fixed speed, one can anticipate the nominal distance a robot is expected to traverse within a set time frame. Yet, challenging terrains, like steep slopes or obstacles, can cause deviations in the actual traveled distance from this estimation. Using the difference between the expected and actual distances within a windowed time period, we can obtain a robust, continuous traversability cost label. For example, while descending a slope, gravity might enable the robot to cover more distance than predicted. Conversely, if the robot encounters an obstacle like a rock and gets stuck, the traversability cost at that region might peak at 1.0, its maximum value. The illustration in \autoref{fig:locomotion} provides a visual representation of this concept, highlighting the disparity between predicted and actual distances.

Additionally, our approach factors in past IMU readings, specifically extracting the acceleration covariance, which serves as neural network input. The nominal distance $d_n$ is computed as:

\begin{equation}
    d_n = ((r.w_1) + (r.w_2)) \cdot t
\end{equation}

Here, $r$ denotes the wheel radius, $w_1$ and $w_2$ represent the left and right wheel speeds and $t$ is the designated period of time. By default, $t$ is 3 seconds unless otherwise specified. Calculating the actual traveled distance, $d_a$, is straightforward, drawing from the robot's state estimation that employs an Extended Kalman Filter (EKF) to integrate IMU, wheel odometer, and GNSS data. The primary purpose of the EKF state estimation is to furnish an initial estimate for ICP-based point cloud mapping. While not strictly essential, the availability of GNSS data could benefit long-term operations, especially during the testing phase. This is because robot movement is relatively restricted in the training phase (three minutes max), making GNSS support advantageous for extended operational periods. 

We maintain label consistency by ensuring that the difference between nominal and actual distances (denoted as $d_n-d_a$) falls within the 0 to 1 range. This is accomplished by clipping values that fall outside this range to their respective minimum or maximum limits, depending on whether they are below the minimum or above the maximum threshold. As our strategy also integrates IMU data, we include the acceleration covariance from the preceding $t$ time and roll, pitch, and yaw angles in the inertial frame. To simplify input for the neural network, orientation is represented using quaternions. Our IMU feature vector is 13-dimensional; 9 elements originate from the flattened acceleration covariance matrix $Q$, while the remaining four belong to the quaternion vector $H$.

\subsection{Environments and Data Generation Process}

\begin{figure}[!htb]
    \centering
    \begin{minipage}{1\textwidth}
        \centering
        \includegraphics[width=1\linewidth]{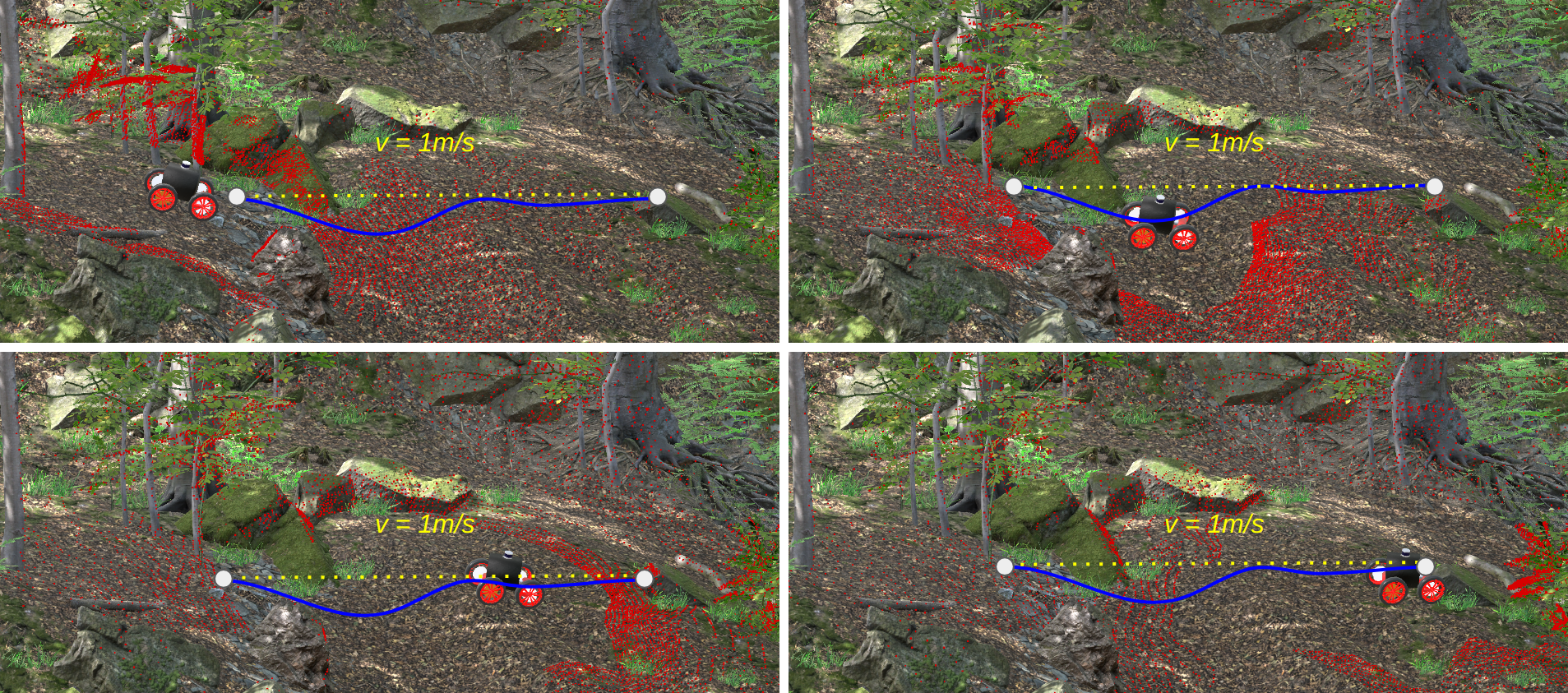} 
        \label{fig:figure1}
    \end{minipage}\hfill
    \caption{
    A sequence of snapshots taken from a high-fidelity forest environment during a robot's data collection journey. Starting at the upper left, the robot progresses to its final position in the bottom right, where it encounters a non-traversable rock, marking the conclusion of a data collection "episode". The locomotion-based traversability, as discussed earlier, is implemented. Data samples are taken at intervals, ensuring the robot covers at least 1 meter between each data snapshot, aligning with the traversability label.}
    \label{fig:data_collection}

\end{figure}

To cater to the data requirements of our neural network, we employ high-fidelity simulations using the Unity Game Engine, renowned for its performance and geometric and visual precision. Our data collection hinges on two primary environments:

\textbf{1.} A forest environment, not based on any real-world location, spanning a $1km^2$ area. It boasts intricate vegetation, diverse terrain, and various objects such as trees, rocks, bridges, rivers, and gravel roads.

\textbf{2.} A real-world replica covering parts of the Norwegian University of Life Sciences campus and adjacent residential areas in As City, Norway. We construct this environment in the Unity Editor using C\# scripts, leveraging elevation data, road structures, and other map objects from the Open Street Map (OSM) dataset. It, too covers a $1km^2$ area.

These detailed environments ensure we gather comprehensive data for both urban and outdoor settings.

The actual data and label collection proceeds in two distinct phases:

First, we deploy multiple robots for efficiency. Each robot is equipped with LIDAR, IMU, Camera, and GNSS sensors, with their data relayed to Robot Operating System (ROS)~\cite{ROS} topics.

Leveraging the gathered sensory data, we implement an Extended Kalman Filter (EKF) as a state estimator, enabling us to generate a robot-centric local point cloud map accurately. This process is elaborated in the subsequent subsection. Using the robot's tracked locomotion quality as a basis for the traversability indicator, we determine the traversability cost label crop points directly below the robot and ascertain the IMU covariance matrix. In the procedure, every sample is stored on the disk with a timestamp, embedding the traversability cost label with three-decimal precision in addition to saving the disk text file including a 13-dimensional IMU feature vector.

Operating multiple robots simultaneously is critical in our data collection methodology. Our experimental setup employs a gaming computer with 32 GB RAM and an NVIDIA 2060 4GB GPU. This configuration permits the concurrent operation of up to 10 robots without any detrimental impact on the frame rate. It is important to recognize that this figure is inherently tied to the computational capabilities of our specific setup, particularly regarding the demands of simulating a 64-ring LIDAR sensor for each robot. 

\begin{figure}[!htb]
    \centering
    \begin{minipage}{1\textwidth}
        \centering
        \includegraphics[width=1\linewidth, trim={0 5cm 0 5cm},clip]{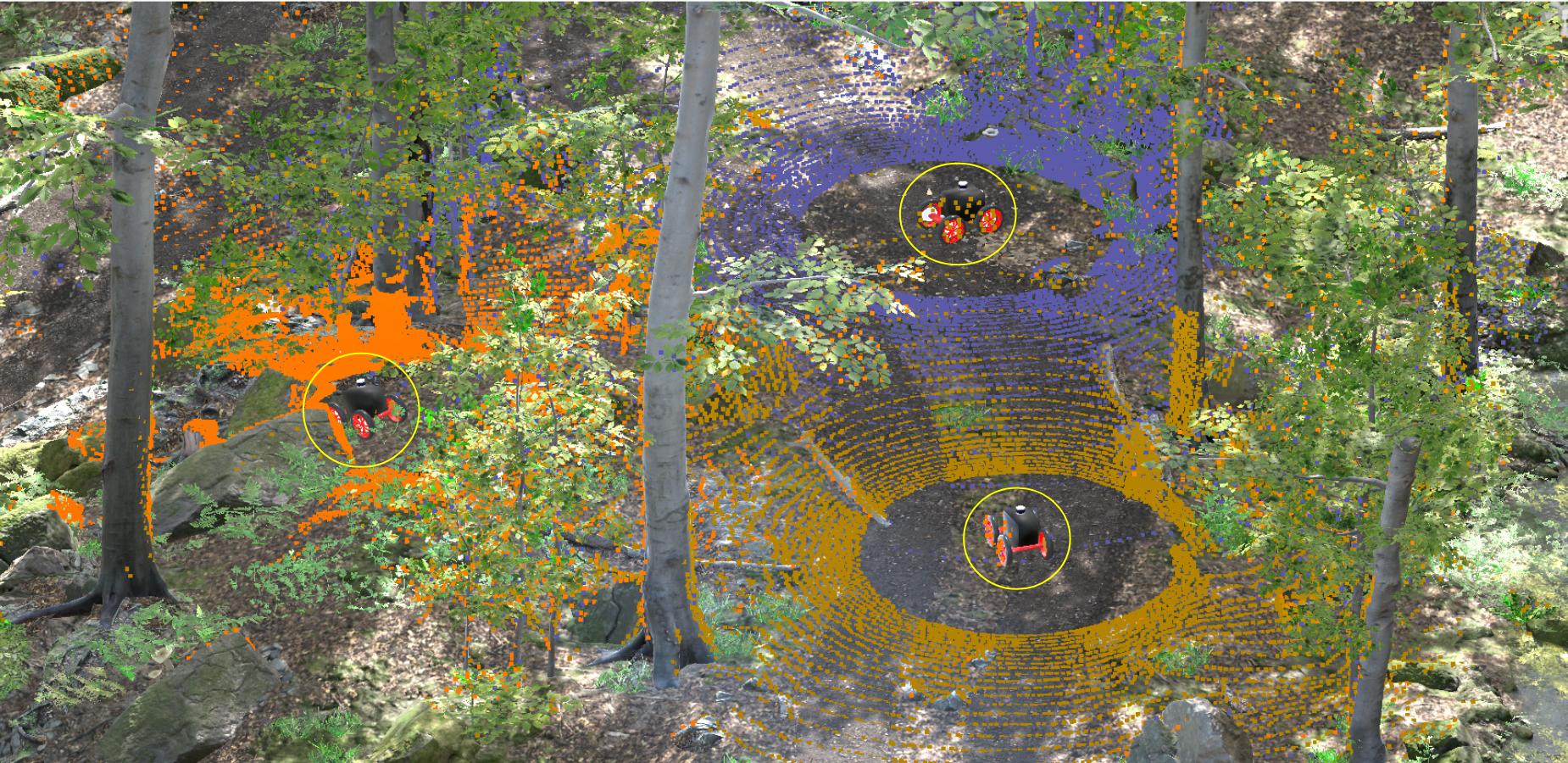} 
        \label{fig:figure1}
    \end{minipage}\hfill
    \caption{
    The image depicts three robots (highlighted within yellow circles) navigating through a simulated high-fidelity forest environment and collecting data alone. The LIDAR point clouds of each robot are also depicted in distinct colors. It is important to note that these robots operate concurrently and can physically interact, with potential collisions and the possibility of one robot's body being detected by the sensors of another. The robots are spawned at random locations with random orientations but move with a fixed linear speed of 1 meter per second throughout each episode.}
    \label{fig:multirobot_collect_data}
\end{figure}

The automated data generation steps are vividly depicted in \autoref{fig:data_collection}, where the anticipatory behaviors of the robots within the simulation are illustrated. Furthermore, \autoref{fig:multirobot_collect_data} expands on this by showcasing the multi-robot data collection process in a life-like simulated environment. These visual representations provide a clear understanding of the sophisticated data generation pipeline that underpins our research.

\subsection{Generation of Dense Robot-Centric Point Cloud Maps}

LIDAR is commonly used to gather precise measurements of the surroundings, though it often provides sparse data compared to the dense information from RGB camera images. A single LIDAR scan might be enough for tasks such as identifying objects or binary categorization of areas as traversable or not. However, when assessing the detailed "traversability" cost of an area with a continuous value, a single scan may fall short. That's why we use point clouds that accumulate data over a windowed time, moving along with the robot, making them robot-centric.

We employ the Iterative Closest Point (ICP) algorithm for generating dense point clouds, benefiting from its efficient GPU performance~\cite{Koide2021}. This method not only improves the accuracy of the robot's state estimation but also effectively tracks movement on steep slopes, creating detailed local point cloud maps. Our approach is also compatible with established SLAM techniques like LIO-SAM (Shan et al.~\cite{liosam2020shan}). However, while these techniques are reliable in many scenarios, they face challenges on highly uneven terrains. For instance, the reliance on IMU pre-integration can lead to misinterpretation of gravitational forces as movement when a robot is stationary on a slope, which is why we did not prioritize this approach.

Loop closures, which help correct the robot's trajectory and avoid drift in traditional SLAM systems, are not crucial for our purpose. This is because our robots generally move in a consistent direction until an obstacle stops them or a set time limit is reached (in the data collection phase). For data collection, we typically let each robot episode run up to a maximum of three minutes to ensure a balanced dataset.

\section{Experiments}
\label{sec:Experiments}

\subsection{Platform, Sensors, and Setup}

\begin{figure}[!htb]
    \centering
    \begin{minipage}{1\textwidth}
        \centering
        \includegraphics[width=1\linewidth]{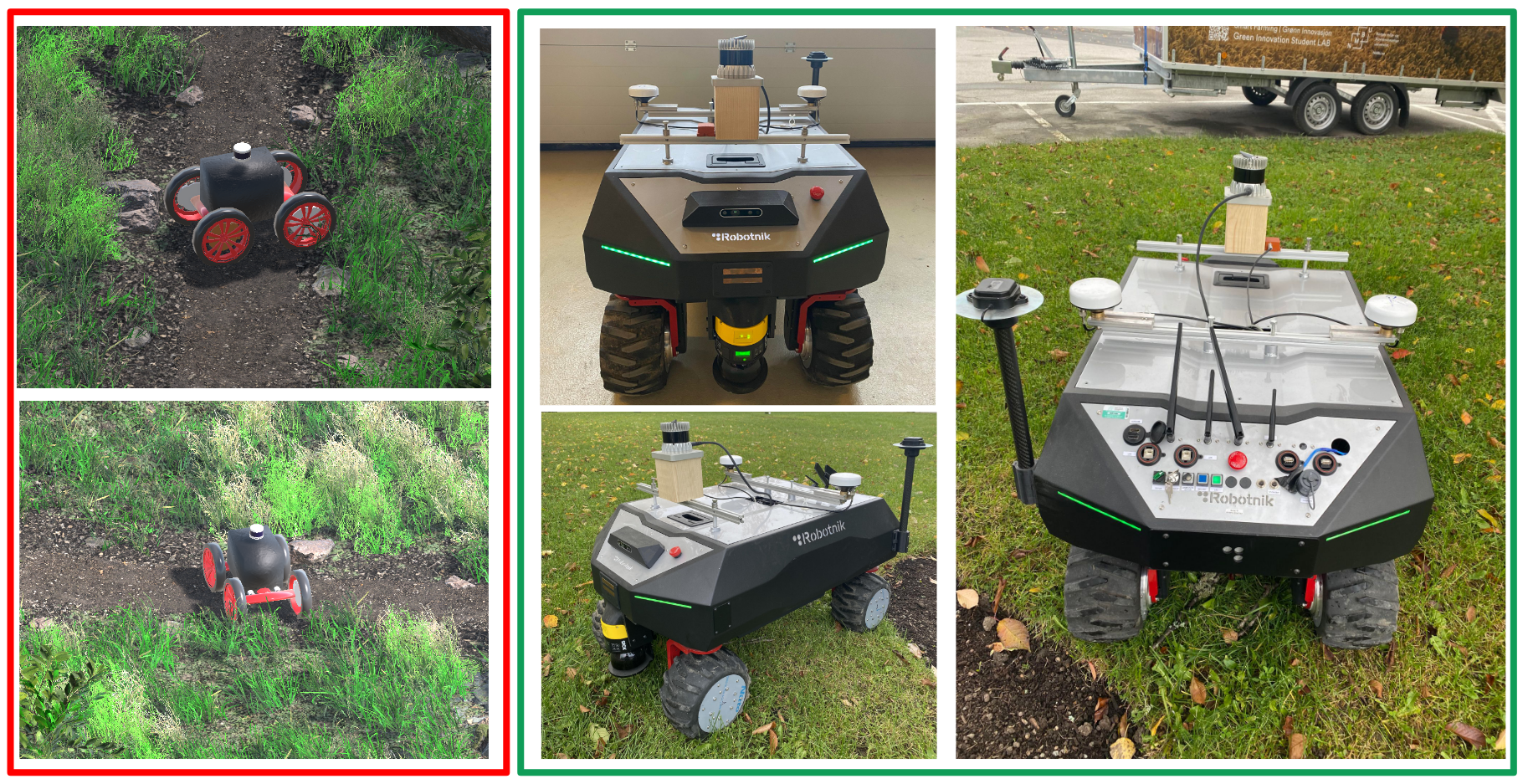} 
        \label{fig:figure1}
    \end{minipage}\hfill
    \caption{Robotic platforms used for experimentation. The red-bordered images depict the simulated robot model, whereas the green-bordered images provide multiple perspectives of the actual Robotnik AGV platform. For our method and experiments, the primary sensors employed are 3D LIDAR and IMU; other sensors on the platform were not directly utilized for traversability estimation.}
    \label{fig:platform}

\end{figure}

The real robot platform we use for real-world experiments is a mid-sized AGV from Robotnik with active-front-and-rear-steering, see \autoref{fig:platform}. To facilitate data collection, we constructed a simulated skid-steered replica of this robot within the Unity game engine. Details of our simulated robot are as follows:

The replica model "Dobbie" is designed to navigate uneven terrains. It is equipped with a 64-ring Ouster LIDAR and an XSENS IMU. The platforms feature additional sensors, including cameras and a GNSS system, though these were not leveraged in developing our traversability estimation framework. While the steering kinematics are not identical, they possess same sensor suite with comparable geometric dimensions and ground clearance. 

\subsection{The Dataset and Data Augmentation}

Our dataset was generated through an automated approach; we created a series of C\# scripts to achieve this. We first position robots at random coordinates within a 1km x 1km environment in Unity. Each robot was oriented randomly and maintained a fixed linear speed of 1 meter per second. The movement continued for either a maximum duration of three minutes or until the robot became immobilized due to obstacles or tipping incidents. We achieved these programmatically by observing the ground truth pose of robots accessible within the simulation.


Throughout the robot's trajectory, our scripts captured snapshots. These comprised a cropped point cloud segment, highlighting the robot's expanded footprint (1.0, 0.67, 1.0) along the x, y, and z dimensions. Accompanying this was the IMU acceleration covariance based on the previous three seconds and the most recent IMU orientation within the inertial frame. Thus, each sample incorporated a point cloud segment of the robot's footprint, the IMU's 3 x 3 covariance matrix, and a quaternion representation of the inertial orientation. The IMU data was simulated based on the rigid body acceleration of the robot.  


To achieve a comprehensive representation of various terrain situations, especially those involving non-traversable regions that are inherently difficult to access, we employed manual teleoperation. This approach facilitated the navigation of robots into such challenging areas. Subsequently, a ROS service was utilized to initiate the capture of samples in these specific locations, ensuring a balanced dataset.
 
To augment the diversity of our dataset, we employed two techniques specifically for augmenting point cloud samples while leaving the IMU data untouched. First, we downsampled the point cloud using a random voxel size selected from the range of 0.05 to 0.25. Second, we introduced zero-mean Gaussian noise with a standard deviation of 0.05 to the point cloud samples. We randomly selected augmentation samples, constituting half of the entire training set. Consequently, post-augmentation, the dataset expanded by a factor of 1.5X. All point cloud samples were subsequently normalized to fit within a metric sphere, while the IMU data remained without any normalization as the 13-dimensional IMU features vector ended up having an intrinsically structured range.

\subsection{Metrics and Evaluation Considerations}

We employ the L1 loss function for our neural network training due to its robustness to outliers. Furthermore, to assess the accuracy of the predicted traversability costs, we utilize the Mean Absolute Error (MAE) metric given by:
\begin{equation}
    \text{MAE} = \frac{1}{n} \sum_{i=1}^{n} |y_i - \hat{y}_i|,
\end{equation}
where $\hat{y}_i$ predicted value, $y_i $ is the actual label and $n$ is number of samples.
We conducted an ablation study to assess the distinct impacts of input features and the incorporation of IMU sensing on our model's overall performance. For example, we included the point normals as additional features to the neural network to see the impact. The performance metrics are presented using MAE values presented in \autoref{tab:training_results}.

\subsection{Feature Impact Analysis on Model Performance}

\begin{table}
\small
\centering
\begin{tabular}{lcccc}
\toprule
\textbf{Features} & \textbf{Final Loss} & \textbf{Train MAE} & \textbf{Test MAE} \\
\midrule
XYZ               & 4.1 & 0.05 & 0.030 \\
XYZ + N.          & 4.0 & 0.05 & 0.035 \\
XYZ + C.          & 4.3 & 0.07 & 0.033 \\
D-F IMU + XYZ + N.    & 4.2 & 0.08 & 0.032 \\
M-F IMU + XYZ         & 3.7 & 0.05 & 0.032 \\
M-F IMU + XYZ + C.    & 3.8 & 0.07 & \textbf{0.024} \\ 
M-F IMU + XYZ + N.    & 3.5 & 0.06 & 0.034 \\
\bottomrule
\end{tabular}
\caption{Training results on automatically generated traversability dataset, the results for different feature combinations and IMU data fusion strategies are presented. The train MAE numbers are peak values during the training. The Test MAE values were retrieved after 300 epochs of training.}
\label{tab:training_results}
\end{table}

In this subsection, we showcase the impact of various combinations of point cloud features and IMU data on the model's performance. Additionally, we report the MAE values derived when evaluating the model on previously unseen samples from the test set.

In \ref{tab:training_results}, we detail the outcomes derived from different feature combinations. Here, "N." signifies normals, and "C." represents the curvature of the point. When utilizing Direct Fusion (D-F), we concatenate the IMU sensing directly with the processed point cloud features as depicted in \autoref{fig:base_traversenet}. In contrast, for Mid-Fusion (M-F) (see, \autoref{fig:spec_base_traversenet}), we first extract features from the IMU data using fully connected layers, then fuse these processed features for subsequent prediction tasks.

\textbf{Discussion:} Analyzing the presented results, it is evident that the model "M-F IMU + XYZ + C" stands out with the final loss of 3.8 and the lowest end MAE of 0.024, making it the most accurate by the conclusion of the training. Interestingly, its performance surpasses even its more feature-rich counterparts, such as "M-F IMU + XYZ + N." This suggests that integrating additional features did not always equate to enhanced performance. For instance, incorporating normals (3-dimensional feature) to the base "XYZ" model results in only marginal improvements or, in some cases, even slight regressions in metrics. The optimal balance between model complexity and performance appears to be achieved with the "M-F IMU + XYZ + C" model, making it a promising choice for the rest of the experiments. Furthermore, the application of Mid-Fusion—entailing additional feature extraction from IMU data prior to concatenation with LIDAR features—appears to enhance the model's performance.

We experimented with L1 and L2 loss functions during the neural network training phase. We observed that L1 loss facilitated better convergence and exhibited greater robustness to outliers within the dataset. For optimization, we utilized the ADAM \cite{Kingma2014AdamAM} algorithm, with a learning rate of 0.001.

\subsection{Qualitative Results of Traversability Estimation}

\begin{figure}[H]
    \centering
    \begin{minipage}{\textwidth}
        \centering
        \includegraphics[width=1\linewidth, trim={0 0 0 10cm},clip]{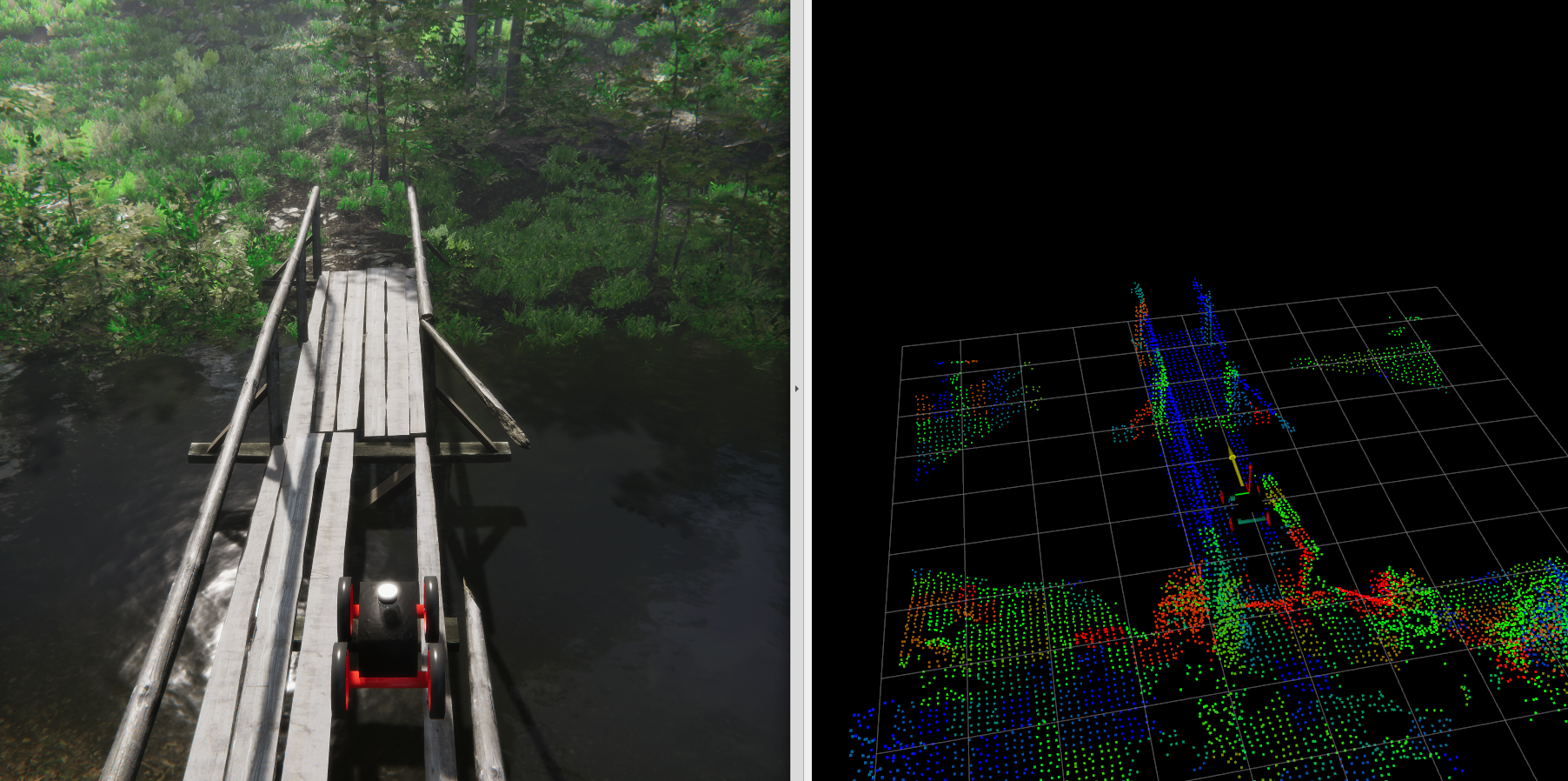}  
        (a)  
        \label{fig:side_by_side_a}        
    \end{minipage}
    \smallbreak
    \begin{minipage}{\textwidth}
        \centering
        \includegraphics[width=1\linewidth, trim={0 0 0 10cm},clip]{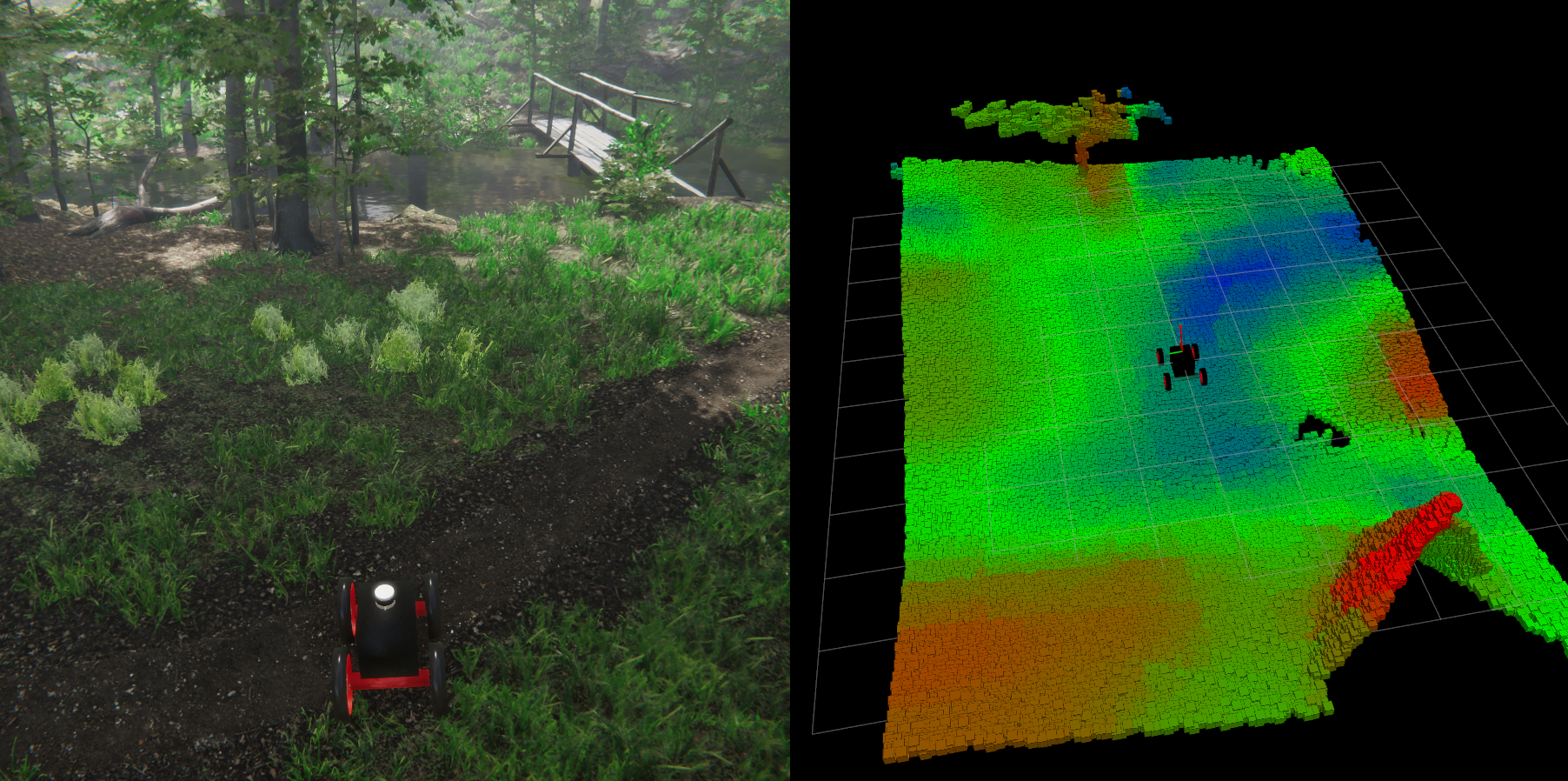}  
        (b)        
        \label{fig:side_by_side_b}
    \end{minipage}
    \smallbreak
    \begin{minipage}{\textwidth}
        \centering
        \includegraphics[width=1\linewidth, trim={0 0cm 0 0cm},clip]{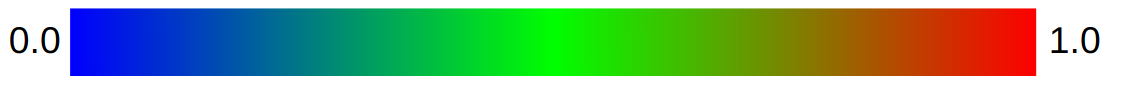}  
        (c)        
        \label{fig:side_by_side_b}
    \end{minipage}
    \caption{The figure illustrates traversability estimations in two distinct environments: a bridge and a random forest patch. In part (a), the robot's local vicinity is discretized to fit within its footprint.
    In part (b), while the environment is again discretized based on the robot's footprint, the inference is conducted on a finer grid resolution. This approach fuses multiple traversability estimations, yielding results that are more accurate and smoother in their presentation. In (c), blue indicates areas that are easily traversable, green suggests moderate challenges, and red denotes difficult or impassable regions, the range within (0, 1).}
    \label{fig:side_by_side}
\end{figure}

\begin{figure}[H]
    \centering
    \includegraphics[width=1\linewidth, trim={0 0cm 0 0cm},clip]{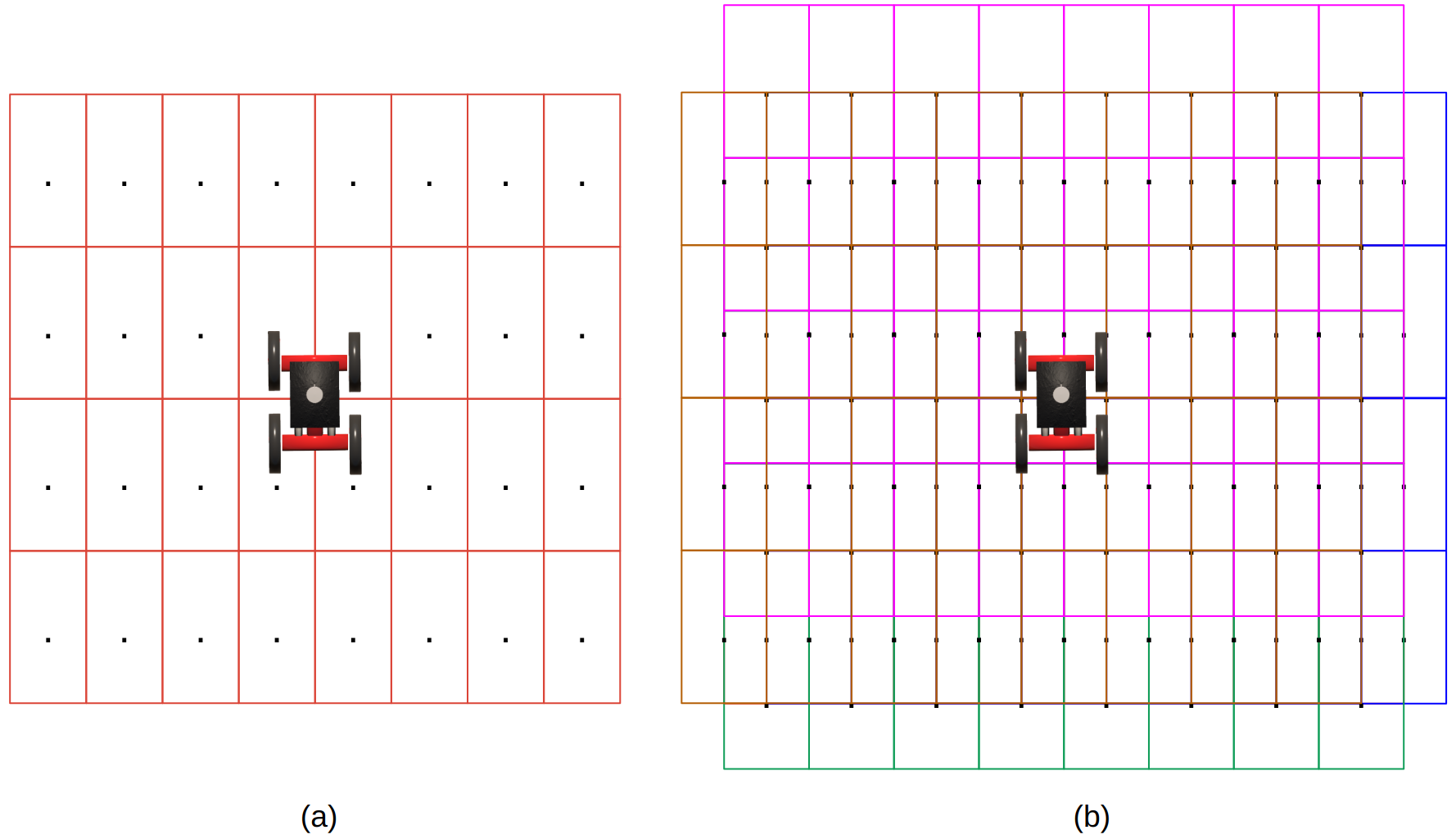}
    \caption{ 
    Two distinct strategies were employed for inferring traversability estimates using the neural network. In the left setup, the environment is discretized based on the robot's footprint size. Conversely, in the right setup, the environment is segmented using the robot's footprint but with a notably smaller step size, resulting in overlapping regions. These overlaps allow for multiple predictions at the same location, enhancing the accuracy and smoothness of the estimates, albeit at the cost of increased computation time.
    }
    \label{fig:grids}
\end{figure}

In this subsection, we detail and interpret the results of our traversability estimation in highly detailed simulation environments. As depicted in \autoref{fig:side_by_side}, the left side visualizes the robot and its environment, while the right side colorfully represents the resulting traversability estimates. When comparing the estimates in \autoref{fig:side_by_side}(a) to those in \autoref{fig:side_by_side}(b), the former appears less smooth. This distinction arises from employing two different inference methods, as further explained in \autoref{fig:grids}. Specifically, \autoref{fig:side_by_side}(b) exhibits a smoother traversability estimate, which aptly allocates lower traversability costs to the more planar regions of the environment. In contrast, while \autoref{fig:side_by_side}(a) shows the robot navigating a narrow bridge and provides associated traversability estimates, its accuracy is not on par with the results in \autoref{fig:side_by_side}(b).

\begin{figure}[htb]
    \centering
    \begin{minipage}{0.48\textwidth}
        \centering
        \includegraphics[width=1\linewidth, trim={0 6cm 0 0cm},clip]{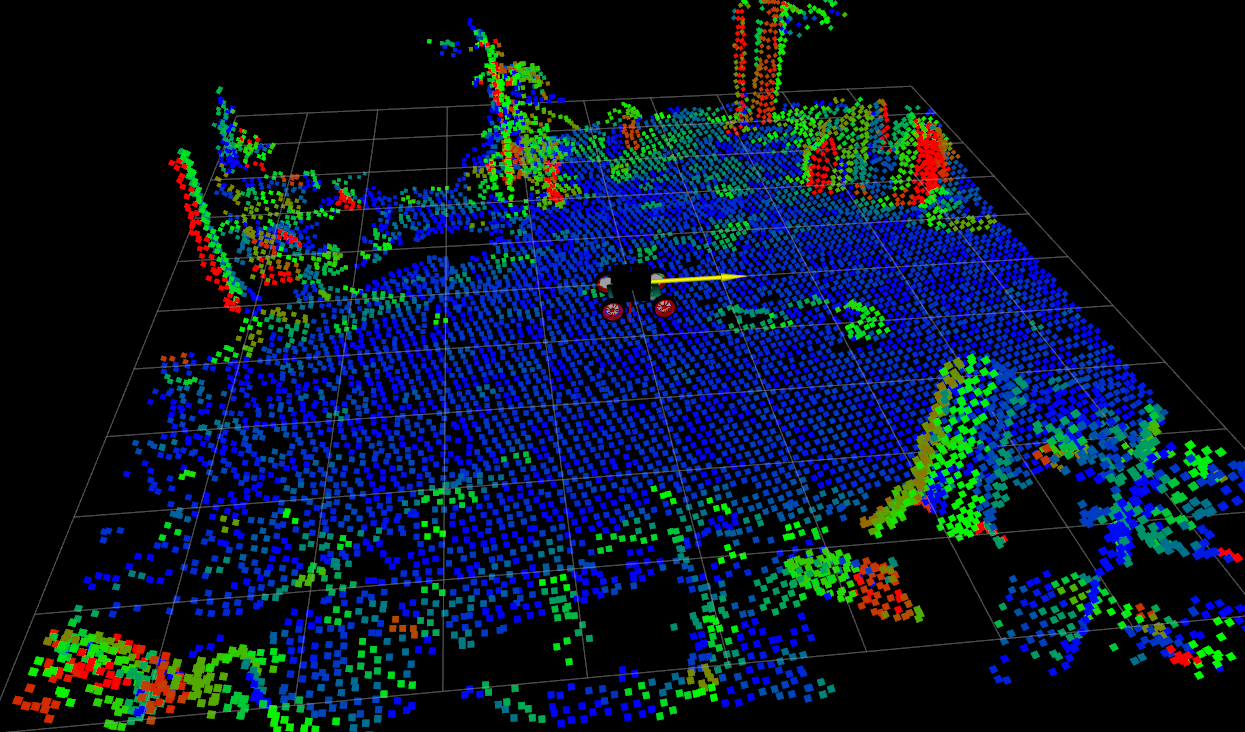} 
        \label{fig:trav_a}
        (a)
    \end{minipage}\hfill
    \begin{minipage}{0.48\textwidth}
        \centering
        \includegraphics[width=1\linewidth, trim={0 6cm 0 0cm},clip]{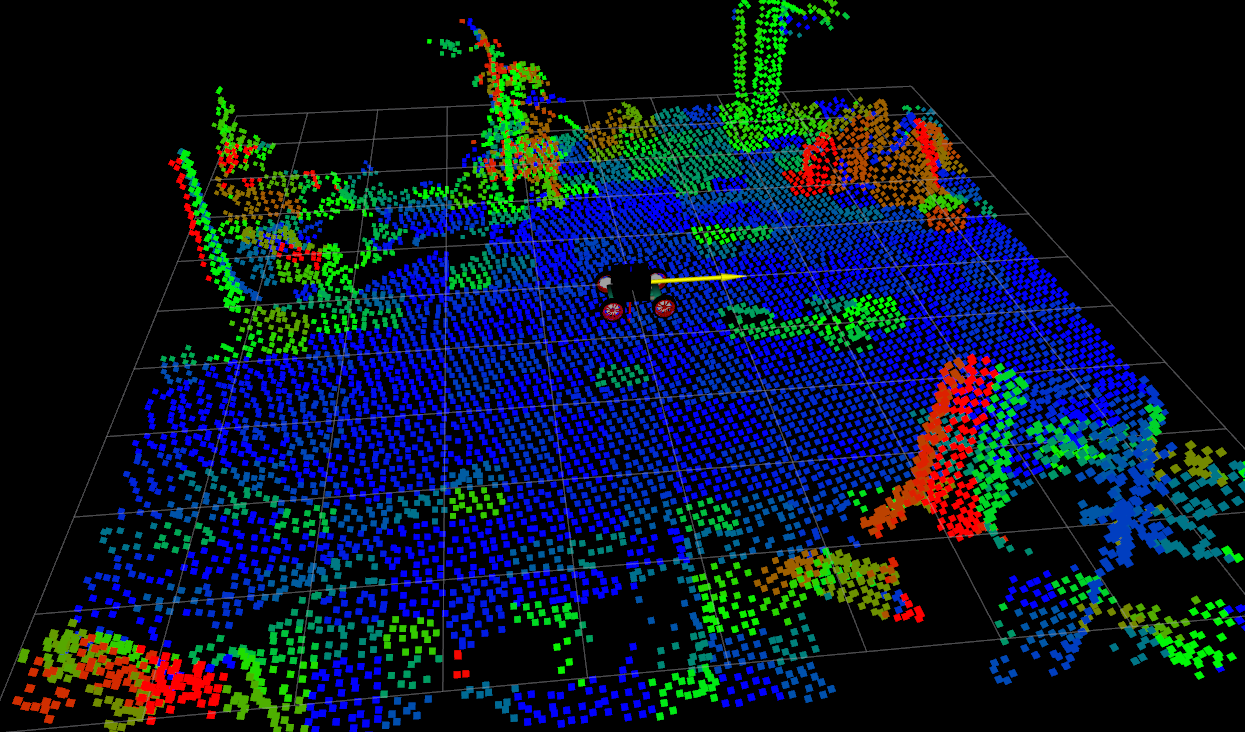} 
        \label{fig:trav_b}
        (b)
    \end{minipage}
    \smallbreak
    \begin{minipage}{0.48\textwidth}
        \centering
        \includegraphics[width=1\linewidth, trim={0 6cm 0 0cm},clip]{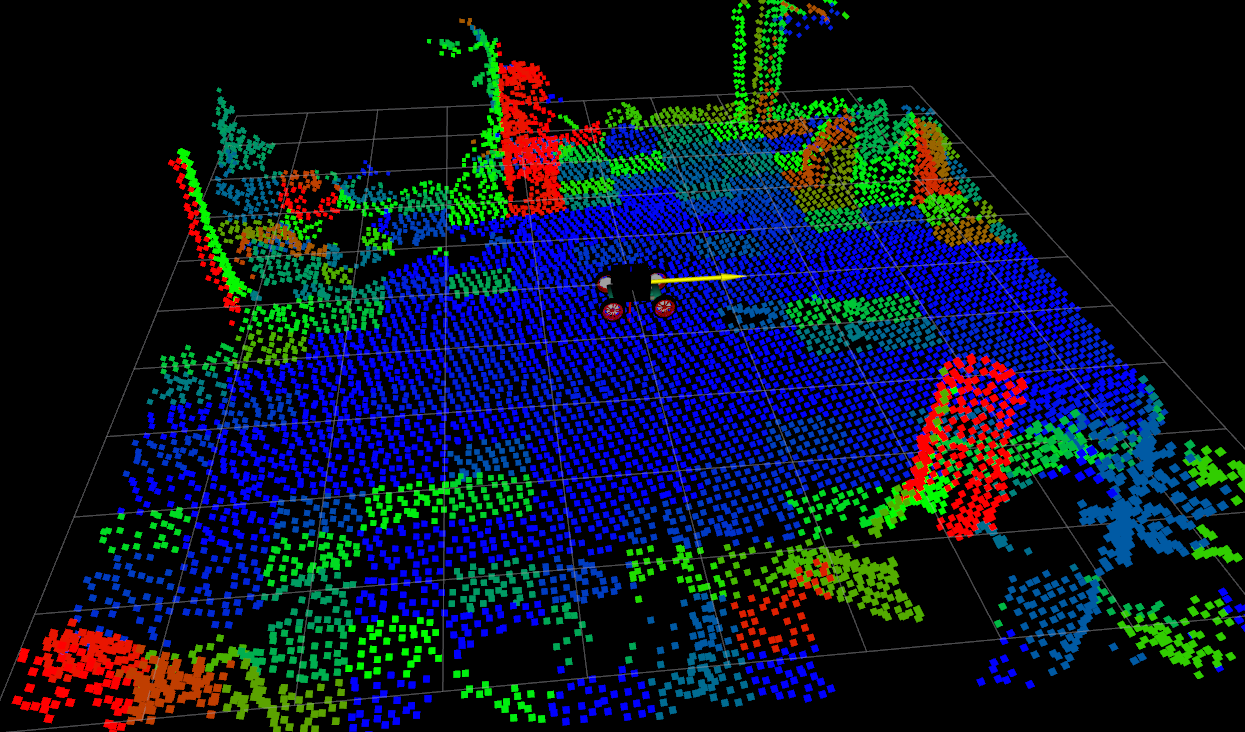} 
        \label{fig:trav_c}
        (c)
    \end{minipage}\hfill
    \begin{minipage}{0.48\textwidth}
        \centering
        \includegraphics[width=1\linewidth, trim={0 6cm 0 0cm},clip]{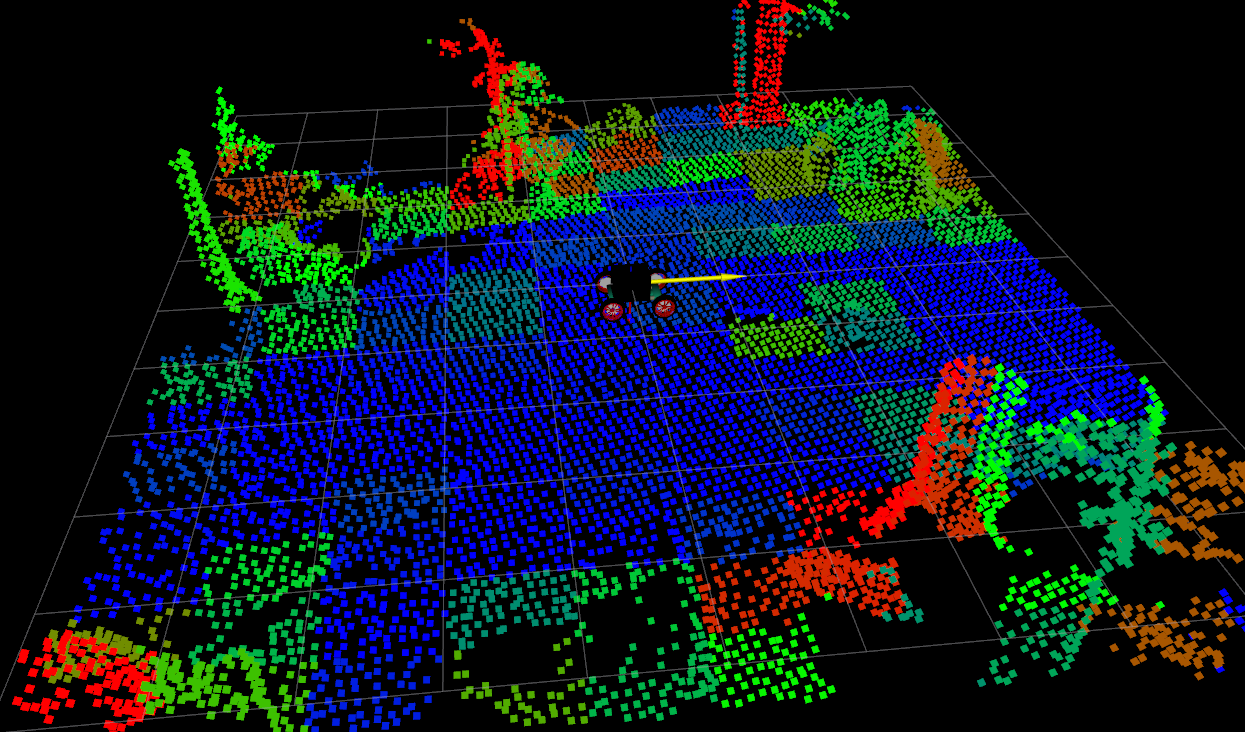} 
        \label{fig:trav_d}
        (d)
    \end{minipage}
    \smallbreak
    \caption{Figures display traversability estimates for robot footprints with a consistent x:y:z ratio as the original footprint of (1.0, 0.67, 1.0). The x dimensions are 0.25 in (a), 0.5 in (b), 0.75 in (c), and 1.0 in (d). }
    \label{fig:different_footprints}
\end{figure}

The size of the robot's footprint dimensions plays a pivotal role in our methodology, as it essentially determines a resolution for the traversability estimation map. In \autoref{fig:different_footprints}, we illustrate the influence of varying footprint sizes on the outcomes. Our proposed neural network architecture was specifically trained with footprint dimensions of (1.0, 0.67, 1.0) for the x, y, and z axes, respectively. Even though optimal results are anticipated when using this specific footprint size, our method remains adept at predicting traversability estimates with changing footprint sizes (given that the ratio of dimensions stays similar). This robustness is attributed to normalizing points within the robot's footprint to a unit sphere during the training and inference stages. In \autoref{fig:different_footprints}(a), we demonstrate the ability to retrieve traversability estimates even with a reduced footprint size of (0.25, 0.16, 0.25). This capability is attributed to our method of aggregating the spatiotemporal LIDAR scan into a dense, robot-centric map that rolls with the robot.

\begin{figure}[htb]
    \centering
    \begin{minipage}{0.48\textwidth}
        \centering
        \includegraphics[width=1\linewidth, trim={0 4cm 0 5cm},clip]{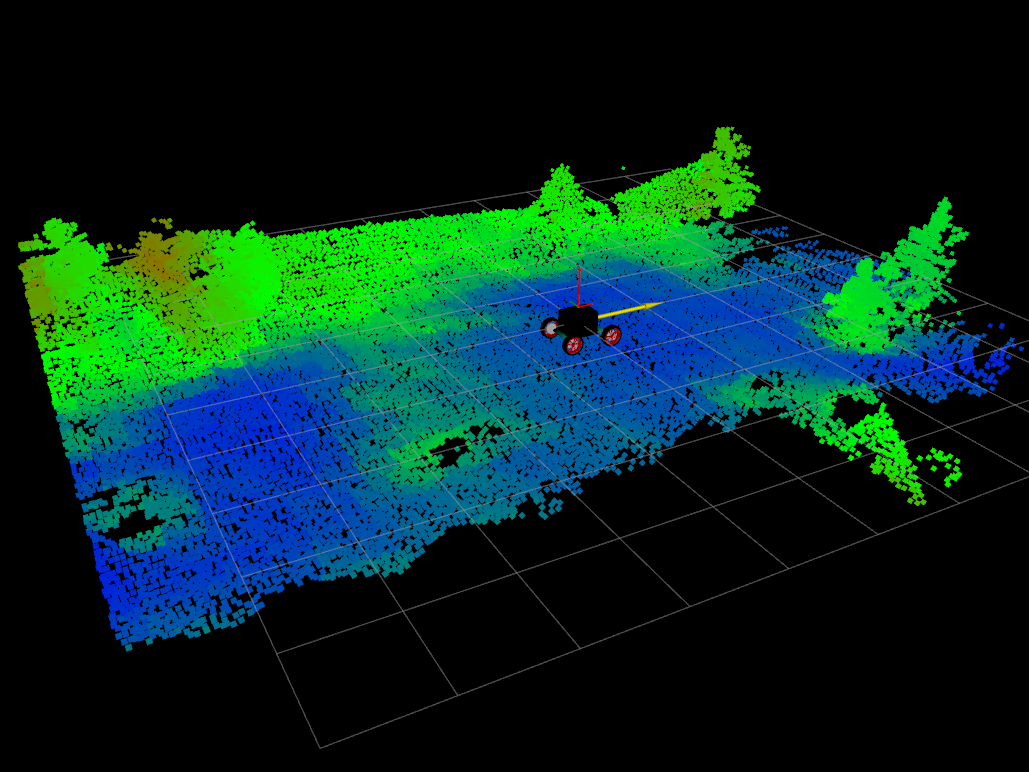} 
        \label{fig:trav_a}
        (a)
    \end{minipage}\hfill
    \begin{minipage}{0.48\textwidth}
        \centering
        \includegraphics[width=1\linewidth, trim={0 4cm 0 5cm},clip]{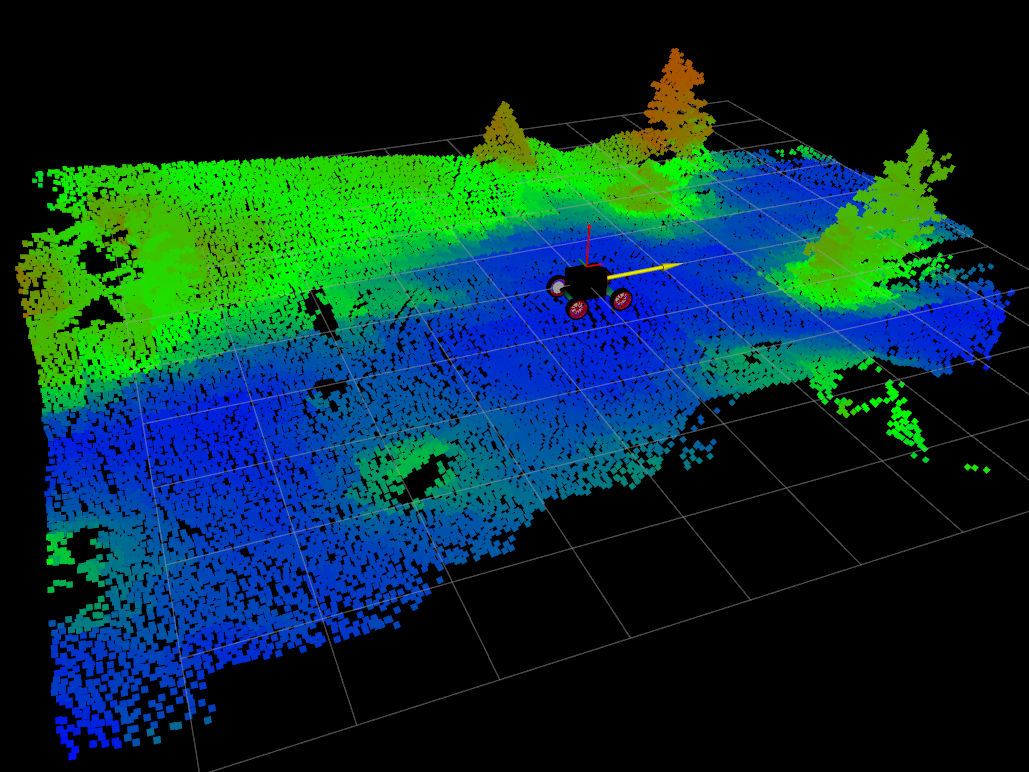} 
        \label{fig:trav_b}
        (b)
    \end{minipage}
    \smallbreak
    \begin{minipage}{0.48\textwidth}
        \centering
        \includegraphics[width=1\linewidth, trim={0 4cm 0 5cm},clip]{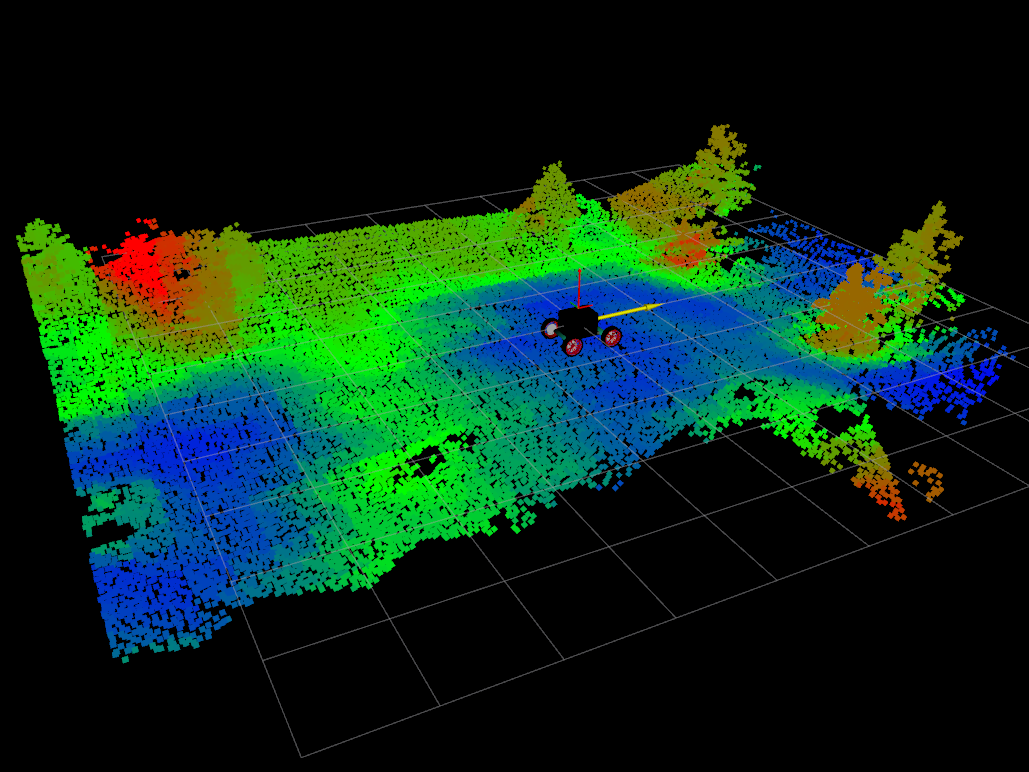} 
        \label{fig:trav_c}
        (c)
    \end{minipage}\hfill
    \begin{minipage}{0.48\textwidth}
        \centering
        \includegraphics[width=1\linewidth, trim={0 4cm 0 5cm},clip]{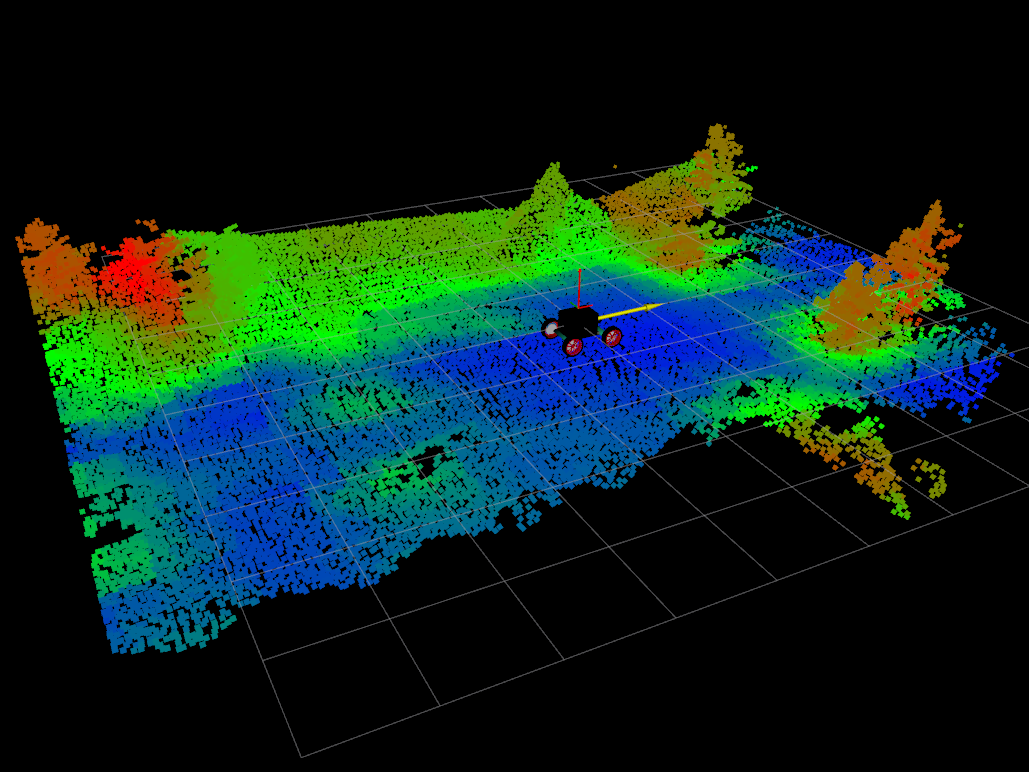} 
        \label{fig:trav_d}
        (d)
    \end{minipage}
    \smallbreak
    \caption{Figures show the effect of robot footprint dimensions on traversability, with x values of 0.25, 0.5, 0.75, and 1.0 in (a) through (d) and a consistent x:y:z ratio. Differing from \autoref{fig:different_footprints}, an overlapping box approach is used, refining the estimate precision.
    }
    \label{fig:prob_different_footprints}
\end{figure}

In \autoref{fig:prob_different_footprints}, we detail the outcomes from the neural network using a fine-grained step size configuration, as depicted on the right side of \autoref{fig:grids}. The results for varying robot footprint sizes reveal that smaller footprints make the method less assertive in pinpointing high-cost areas, possibly due to the limited point information and variation in these smaller regions. As the footprint size expands, the method exhibits heightened confidence in distinguishing clearly non-traversable zones, particularly evident in \autoref{fig:prob_different_footprints}(d).

\begin{figure}[H]
    \centering
    \includegraphics[width=1\linewidth]{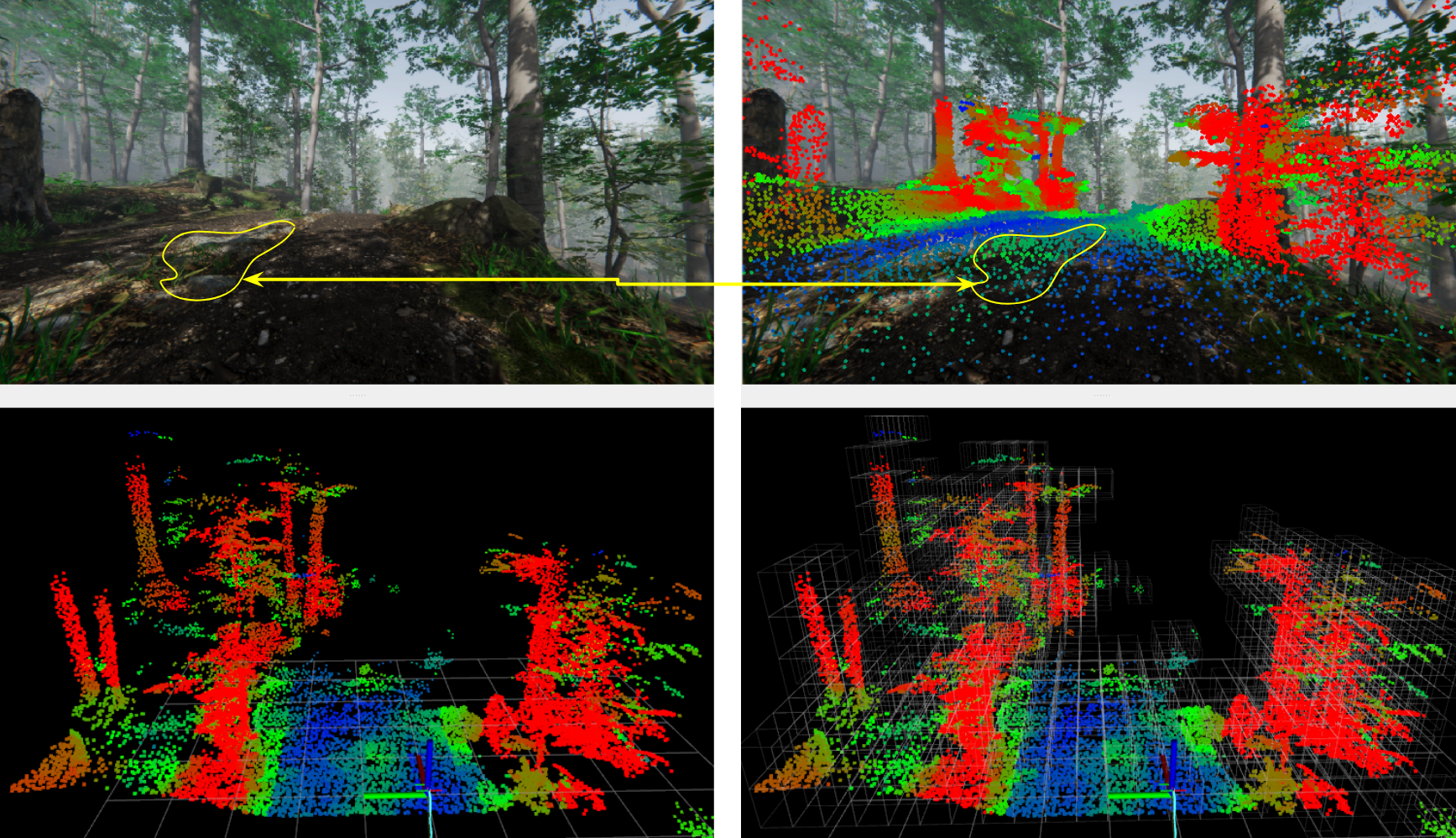}
    \caption{TraverseNet Evaluation in High-Fidelity Simulation: The top-left image presents the robot's view, while the top-right shows traversability estimations overlayed onto this view, allowing for method accuracy verification. The bottom images display the 3D traversability estimation point clouds, with discretized environmental boxes highlighting spatial delineations. } 
    \label{fig:lifelike_sim}
\end{figure}

\autoref{fig:lifelike_sim} showcases additional experimental outcomes from deploying the trained neural network in high-fidelity simulations, illustrating the efficacy of LIDAR in low-light and shadow-prone forested terrains. In the top-right section, the network's capability to discern the 'rough' segments of a gravel path is evident, where higher yet navigable regions are marked in distinct colors upon the rocks, encapsulated by a yellow contour for emphasis and comparative analysis.

\textbf{Discussion:} Our method's proficiency in handling diverse scenes directly, without the need for additional tuning or real-world data, stands out as a key finding in our qualitative assessment. The method employs densified point cloud maps that facilitate detailed traversability estimates, proving especially beneficial for challenging terrain textures and navigating low-light conditions. Moreover, the introduction of overlapping boxes in our approach has proven to refine coarse estimates, resulting in smoother estimates that are favored in path planning and robotic control systems.

Despite the above strengths, our method is not without limitations. It involves certain parameters, such as the size of the time window for IMU covariance computation and the set linear velocity, which are not trivial to optimize. Alterations to these parameters necessitate a re-collection of data, posing a challenge to the method's adaptability.

\subsection{Comparision to state-of-the-art}

\begin{table}[h]
    \scriptsize
    \centering
    \begin{tabular}{|c|c|c|c|c|c|}
        \hline
        \textbf{Method} & \textbf{Data Input} & \textbf{Formulation} & \textbf{Metric} & \textbf{Performance}  & \textbf{Directionality}\\
        \hline
        Our &   P. Clouds & Regression & MAE & 0.024-0.035 & Y \\
        \hline
        Wallin et al.~\cite{Wallin2022} &  DEM & Regression & MAE & $\approx$0.2 & Y \\
        \hline
        Xue et al.~\cite{Xue2023} &   P. Clouds & Classification & Precision & ~97-99\% & N \\
        \hline
        Agishev et al.~\cite{Agishev2023} &   P. Clouds & Classification & - & - & N \\
        \hline
    \end{tabular}
    \caption{Comparative Analysis of Various Traversability Estimation Methods and Their Corresponding Performance Metrics: This table summarizes the performance values of each method as reported in their respective studies, evaluated on the datasets used in those papers. }
    \label{tab:method_comparison}
\end{table}

We have drawn a comparison between our method and that of Wallin et al.~\cite{Wallin2022}. Both techniques gauge traversability by assessing the disparity between the actual traveled distance and the anticipated distance predicated on robot motion parameters. Our method directly processes point clouds, in contrast to Wallin et al., who employ a discrete elevation map. Their reported Mean Absolute Error (MAE) stands at 0.2 for a 1.0-meter grid resolution. In contrast, our findings, detailed in \autoref{tab:training_results}, indicate an MAE of 0.024, marking a considerable enhancement in accuracy for the same resolution.

Furthermore, we evaluated the outcomes achieved by Xue et al.~\cite{Xue2023}. This method, which doesn't rely on learning, operates on spatiotemporally aligned point clouds similar to ours. A crucial distinction is that Xue et al.~\cite{Xue2023} evaluate traversability estimation as a binary classification task. They quantify their metric, termed "precision", as the ratio of true positives to the sum of true and false positives. The precision they report falls within the 97-99\% range.

To evaluate the efficacy of our approach, we further evaluated our method on an external dataset of real-world forest environment data provided by Agishev et al.~\cite{Agishev2023}, which is discussed in the subsequent section.

\begin{figure}[H]
    \centering
    \begin{minipage}{\textwidth}
        \centering
        \includegraphics[width=1\linewidth, trim={0 0 0 0},clip]{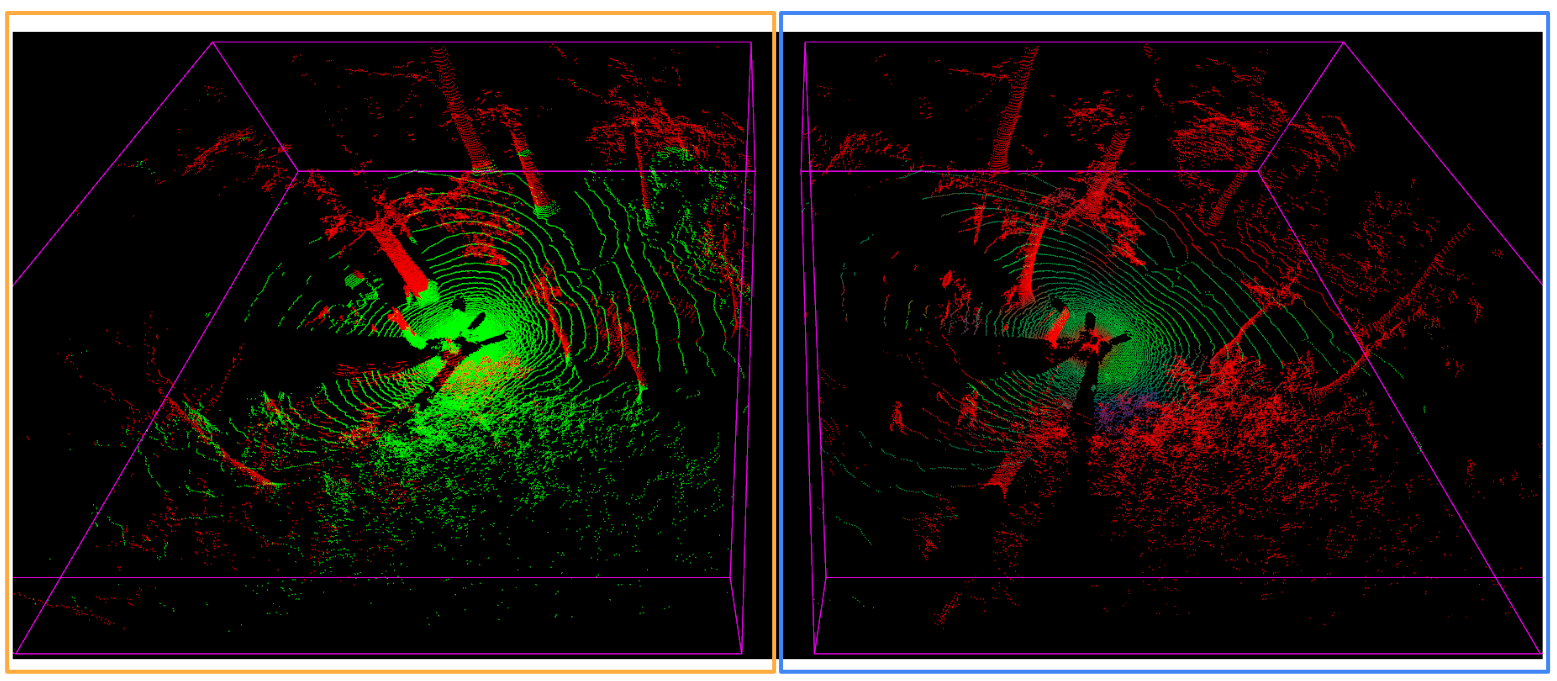}      
    \end{minipage}
    \begin{minipage}{\textwidth}
        \centering
        \includegraphics[width=1\linewidth, trim={0 0 0 0},clip]{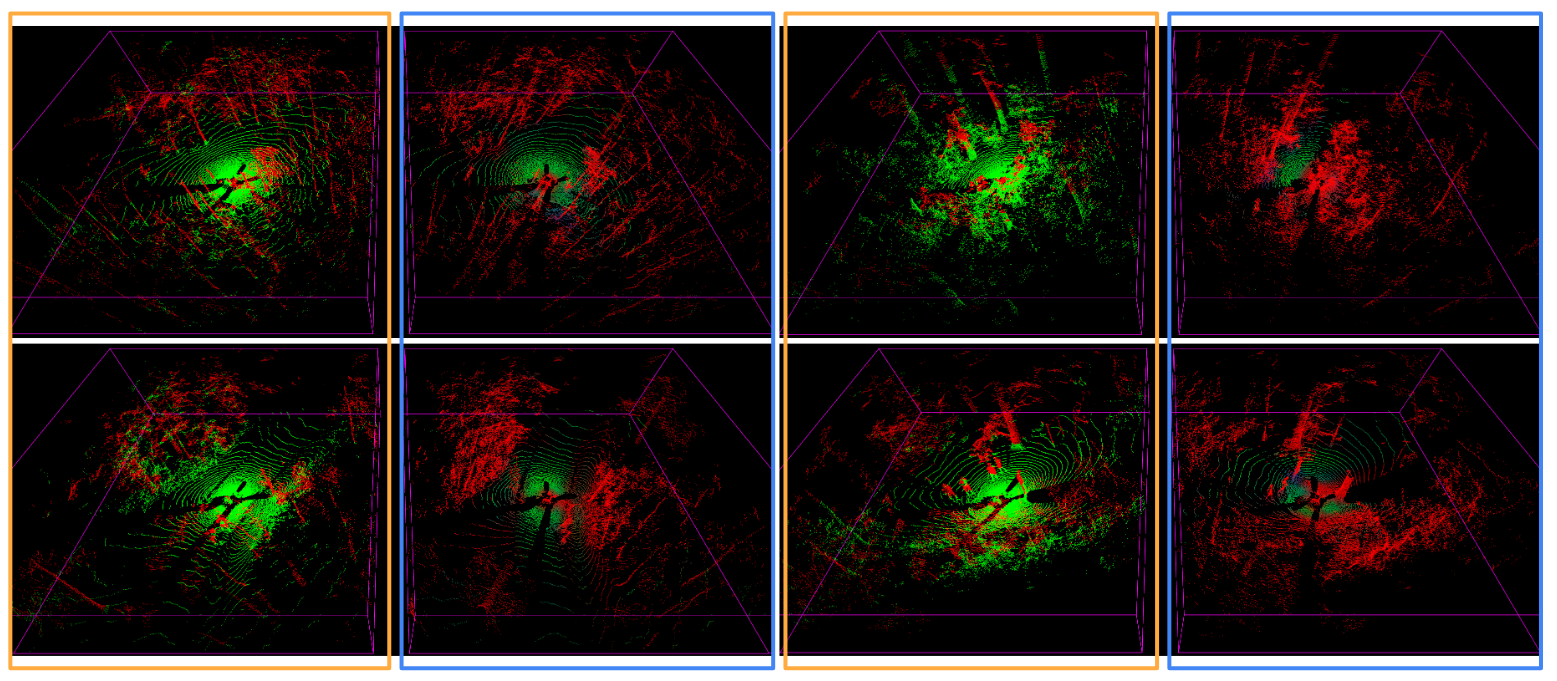}  
    \end{minipage}
    \caption{The figures showcase the efficacy of our traversability estimation method when applied to an external forest dataset provided by Agishev et al.~\cite{Agishev2023}. Our approach yields enhanced accuracy even without IMU data integration. This improvement is especially evident in resolving coarse binary classifications of objects like tree trunks and pedestrians, where our method offers more accurate traversability assessments. Visual comparisons are provided by highlighting our method's estimates in blue boxes, contrasting with the orange boxes used for Agishev et al.'s estimates. }
    \label{fig:cvut_dataset}
\end{figure}

The ambiguous nature of traversability estimation, influenced by factors such as sensor types, varying problem formulations (classification or regression), and the capabilities of different robot platforms, presents a challenge in establishing a standardized benchmark for comparison with the leading methods. However, the dataset released by the authors of \cite{Agishev2023} serves as a suitable candidate for benchmark evaluations. The robot platform they used in the experiments poses locomotion capabilities similar to our robot model. In adapting our approach for this comparison, we adjust our method to yield values aligning with their model's binary output. It is important to note that the dataset includes labels generated by Agishev et al.'s\cite{Agishev2023} method rather than manually verified ground truth. Therefore, the comparative results shown in \autoref{fig:cvut_dataset} are intended for qualitative analysis rather than quantitative validation, as there are no ground truth labels for quantifying both method's accuracy.  

\textbf{Discussion:} Utilizing point clouds offers a distinct advantage due to their ability to capture dense and raw measurements, especially when they are aligned over time using techniques like ICP or NDT, as seen in both our method and that of Xue et al.~\cite{Xue2023}. Unlike elevation maps, which provide a simplified view, point clouds capture the full geometric intricacies. This detailed representation might be a significant factor in the enhanced accuracy demonstrated in \autoref{tab:method_comparison}. Although the results from Xue et al.~\cite{Xue2023} are noteworthy, it is important to highlight that their method views traversability in simpler terms: it is either traversable or not. Moreover, they do not take into account the robot's direction. This distinction is vital; for example, the challenge of moving down a slope with a 10\% grade is different from moving up the same slope. 
The reported performance values are derived directly from the authors' conclusions as stated in their respective sections. Each method underwent evaluation on its own distinct dataset; the absence of accompanying implementations and different sensor modalities constituted a barrier to the replication of the reported performance outcomes.

Upon analyzing the results from the external dataset by \cite{Agishev2023}, our method demonstrates a more cautious and detailed performance. For example, regions at the base of tree trunks and portions of the pedestrian point clouds, which were often marked as traversable by \cite{Agishev2023}, were correctly identified as non-traversable by our method. It is noteworthy, however, that our method's conservative bias may lead to mistakenly categorizing steep slopes as non-traversable, as depicted by the images in \autoref{fig:cvut_dataset}.

\subsection{Real-world Experiments}

\begin{figure}[htb]
    \centering
    \includegraphics[width=1\linewidth, trim={0 2cm 0 2cm},clip]{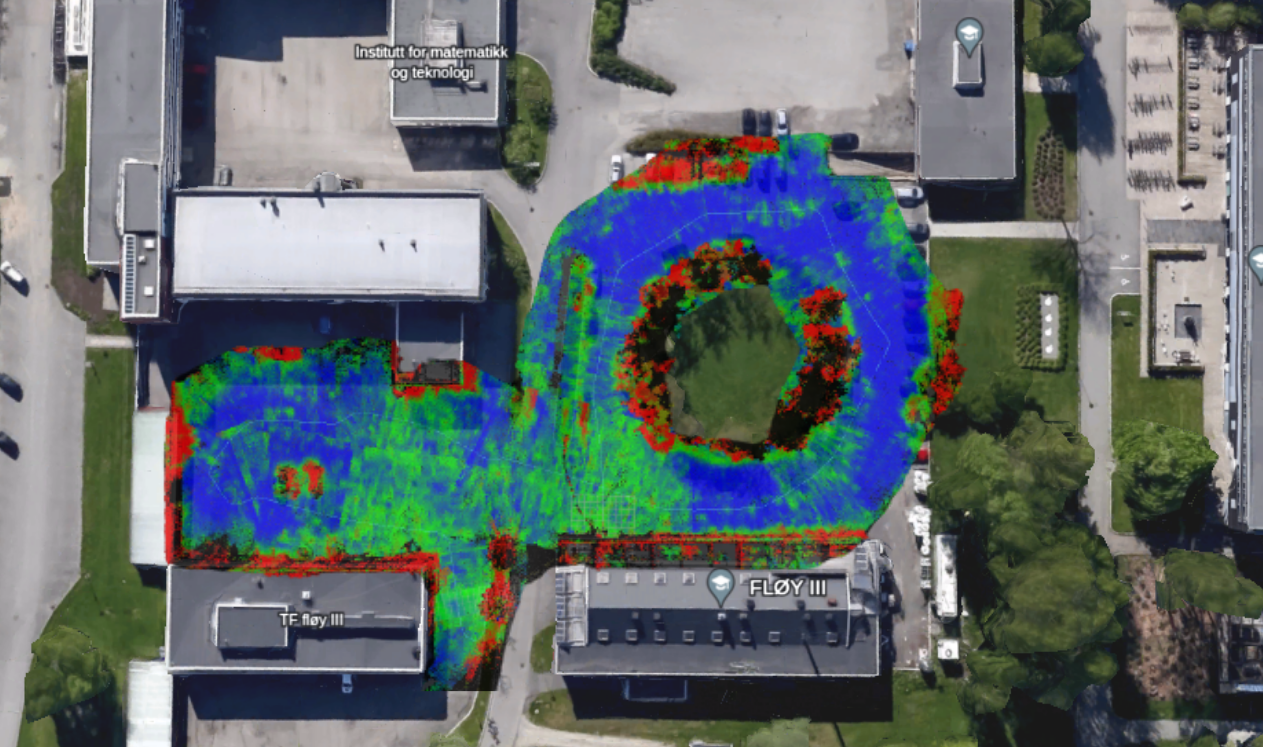}
    \caption{ 
    The traversability estimates are merged with the point cloud map generated using LIO-SAM \cite{liosam2020shan}, a 3D SLAM approach. The method accurately differentiates between solid obstacles like trees and buildings. Additionally, it designates higher traversability costs to areas with inclines or gravel, as indicated by the green patches). The integrated traversability map is overlayed onto Google Earth's satellite imagery for reference.
    }
    \label{fig:real_experiment}
\end{figure}

In our study, we performed real-world experiments using the Robotnik platform depicted in \autoref{fig:platform}. Despite training the neural network exclusively on simulated data, we observed impressive generalization during field tests; the model adeptly estimated traversability without requiring additional fine-tuning. We conducted these tests on the campus of the Norwegian University of Life Sciences, an outdoor setting featuring varied elevations, and sections of asphalt and gravel roads. The derived traversability estimates, integrated into the environmental map, are overlayed onto a satellite image in \autoref{fig:real_experiment}.

Our real experiments indicate that our method can produce a locally consistent, dense traversability map within acceptable time frames. We trim the SLAM-generated map to focus on the robot's immediate vicinity using a 3D box defined by corners (-10, -10, -5) and (10, 10, 5). Using the overlapping box configuration shown on the right of \autoref{fig:grids}, we generate around 200 boxes, discarding those boxes encapsulating fewer than three points. The neural network processes this in approximately 0.2 seconds. Using the non-overlapping box configuration depicted on the left of \autoref{fig:grids} can enhance processing speed to 50 FPS, though at the expense of estimation accuracy. 

\begin{figure}[htb]
    \centering
    \includegraphics[width=1\linewidth]{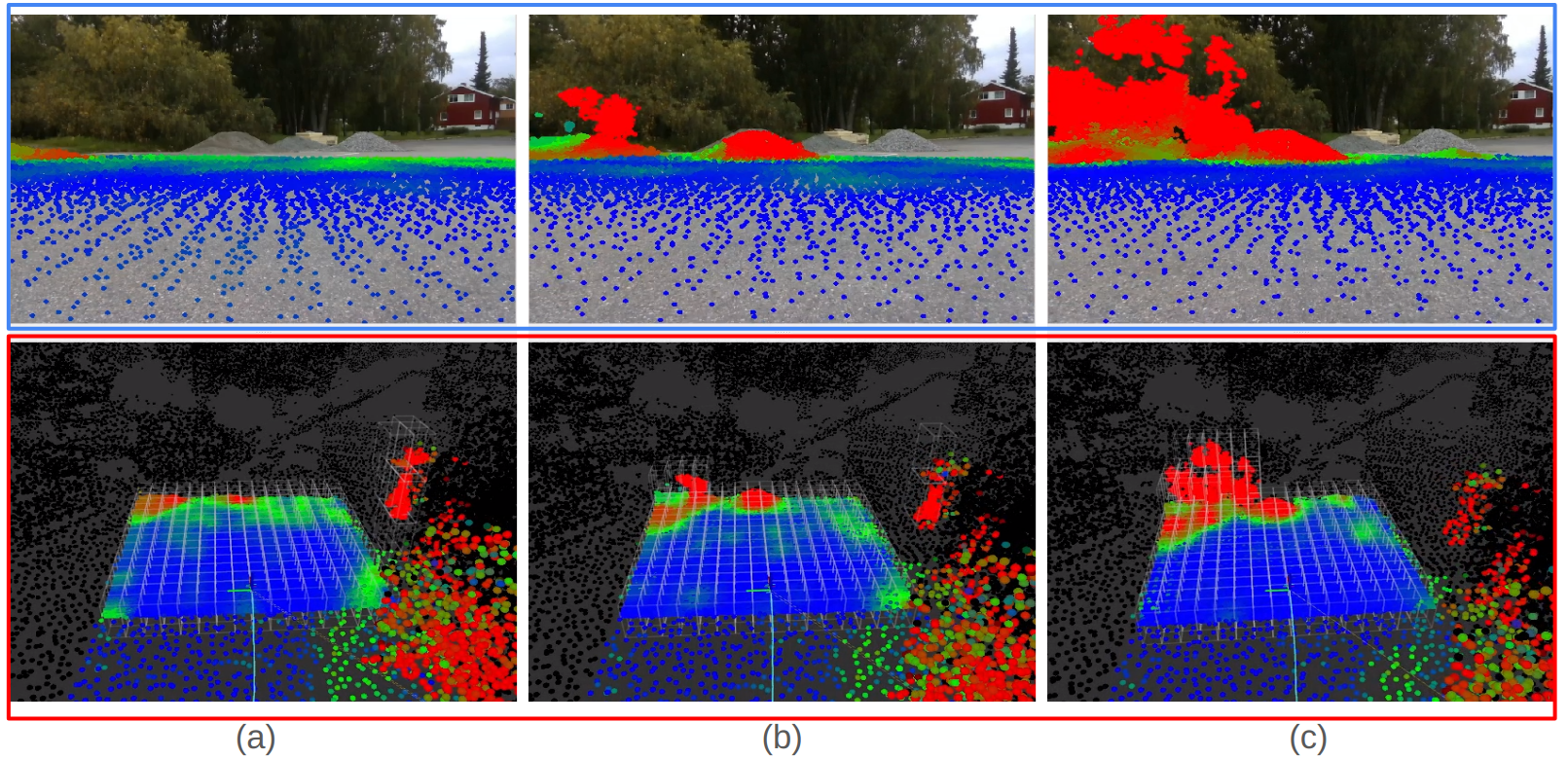}
    \caption{ 
   A sequence showcasing the method applied to a real scene, featuring both image-point cloud views. The top images overlay the traversability point cloud map on actual images for visual validation. The bottom images present 3D visualizations of the traversability estimation within the point cloud, where traversability costs are embedded in RGB values.
    }
    \label{fig:real_experiment2}
\end{figure}

\begin{figure}[H]
    \centering
    \includegraphics[width=1\linewidth]{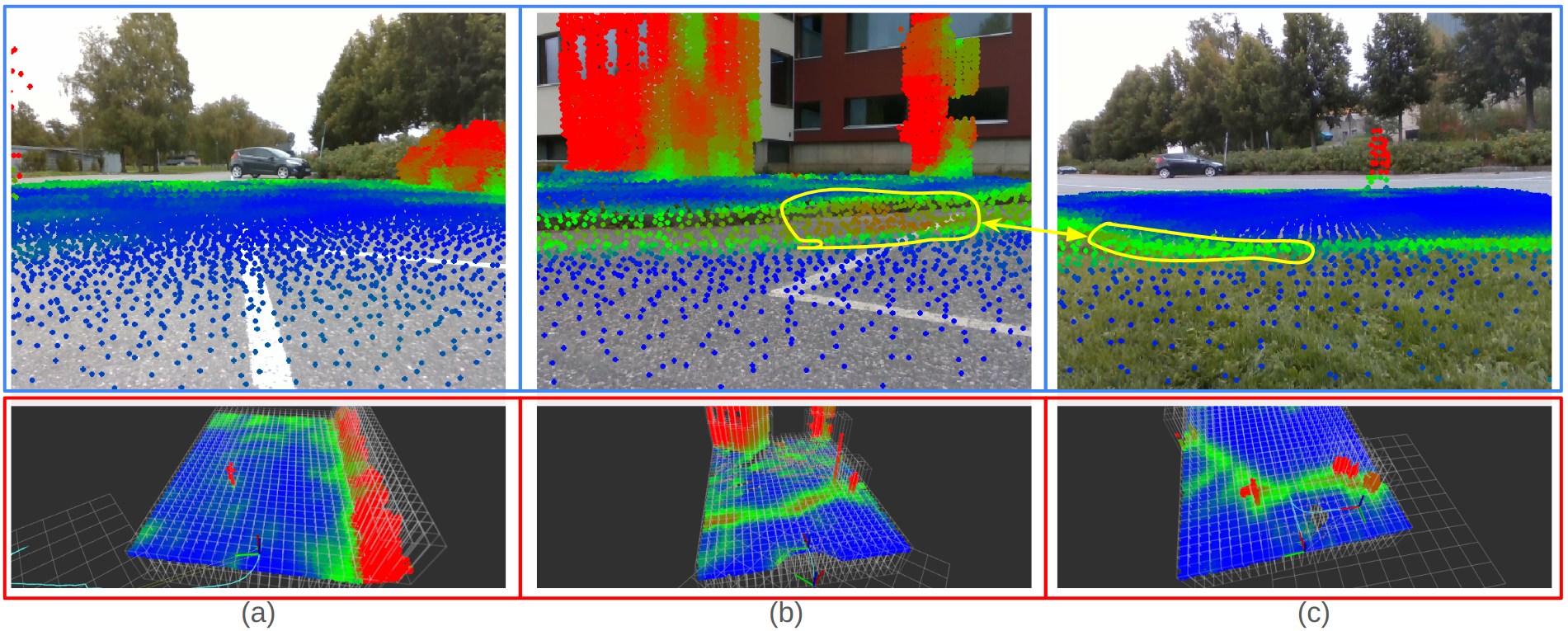}
    \caption{  
    Snapshots from various moments during an outdoor experiment. As in \autoref{fig:real_experiment2}, the image-point pairs are provided where RGB colors of points indicate traversability costs. The figure also demonstrates the method's directionality awareness, assigning higher traversability costs and a larger area of concern when approaching an uphill sidewalk bump (b), as opposed to a less conservative estimate when descending from the same sidewalk (c). The relevant regions have been marked with yellow contours in images (b) and (c) for clarity.
    }
    \label{fig:real_experiment3}
\end{figure}

\textbf{Discussion:}
Our method effectively identifies areas with subtle sidewalk bumps, fences, and tall vegetation, as illustrated in \autoref{fig:real_experiment3}. Moreover, the distinction between gravel and asphalt roads in terms of traversability estimates is apparent in \autoref{fig:real_experiment}, likely influenced by the inclusion of IMU data. However, the method encounters challenges with dynamic obstacles. For instance, in \autoref{fig:real_experiment3}, a pedestrian, once captured in the map but later moved, remains depicted in the traversability estimate. 

\subsection{Path Planning and Navigation Experiments}

\begin{figure}[H]
    \centering
    \includegraphics[width=1\linewidth]{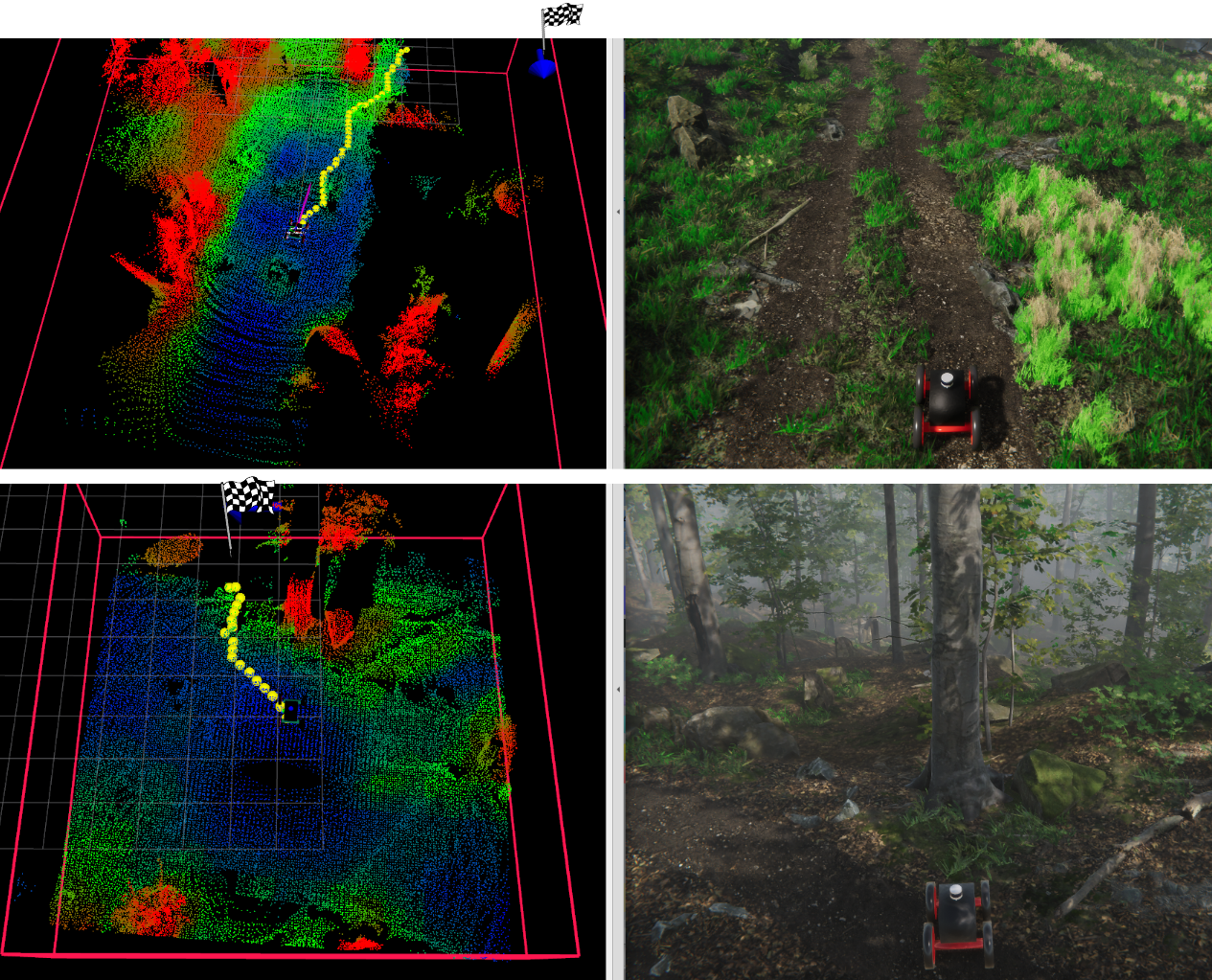}
    \caption{ 
     Snapshots capture two instances where the point cloud-based path planner by Atas et al.~\cite{Atas2023} utilizes our proposed traversability mapping to generate collision-free, cost-effective paths. The traversability map with the optimal path in yellow is shown on the left. At the same time, the right showcases the high-fidelity simulated environment and the corresponding robot positions for the scene.
    }
    \label{fig:planning0}
\end{figure}

We conducted path planning and navigation experiments to evaluate our method in a practical scenario. Planning algorithms utilized the traversability maps to optimize local paths by favoring regions with lower traversal costs in these tests. We employed a variant of the A* algorithm, proposed by Atas et al.~\cite{Atas2023}, which is particularly adept at handling point clouds. Using our traversability estimates, this algorithm refines a prior global plan within the robot's immediate vicinity (highlighted by the red bounding box in \autoref{fig:planning0}).

\begin{figure}[H]
    \centering
    \includegraphics[width=1\linewidth]{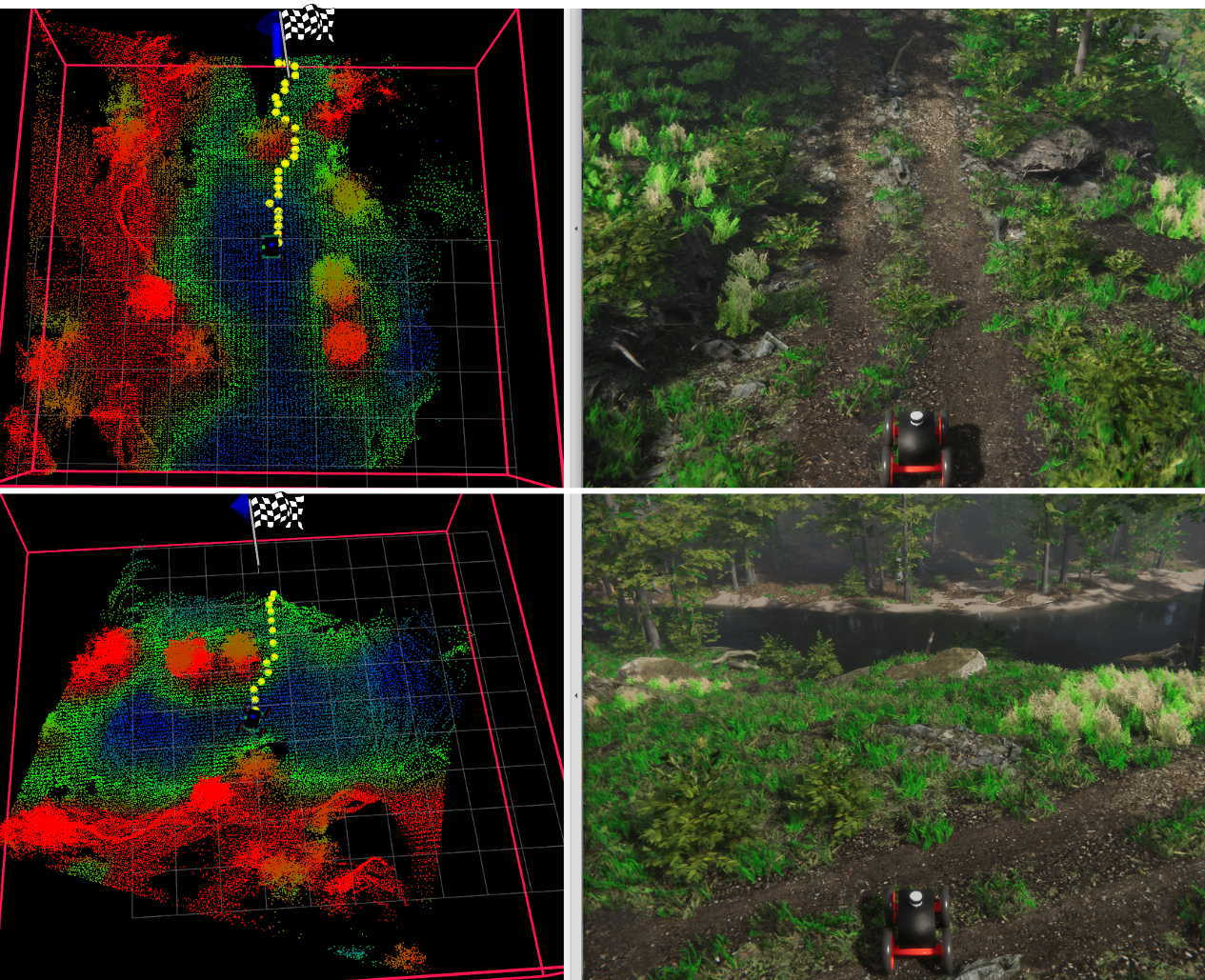}
    \caption{  
     Similar to \autoref{fig:planning0}, the left side displays the traversability cost map with the optimal path highlighted in yellow, and the right depicts the high-fidelity simulated environment and the robot.
    }
    \label{fig:planning1}
\end{figure}

Our experiments further demonstrated an additional application of the proposed method: exploration in the absence of a pre-existing global map or plan. Here, we set a dynamic goal that shifts with the robot's position, enabling it to explore the terrain autonomously until it assesses that further movement may result in a collision. These experiments demonstrate the capability of the proposed method to guide a robot in choosing safe and feasible paths within complex environments.

\section{Concluding Remarks}

This paper introduces a novel neural network architecture to predict continuous traversability values using high-resolution, robot-centric point cloud maps. Leveraging deep neural networks, our method circumvents the tedious process of manual data labeling, as both data and labels are autonomously generated within a high-fidelity simulation environment.

Our extensive experiments validate the robust performance of our proposed method, exhibiting highly accurate predictions on our collected validation and third-party datasets. Furthermore, the pre-trained neural network model was directly deployed on an actual robot platform, providing satisfactory traversability estimates even without integrating any real-world data samples.

The practicality of our approach is further demonstrated through its application in 3D path planning algorithms that operate directly on point clouds. In such applications, our method consistently enabled planners to produce collision-free paths, proving critical for safe navigation and exploration within complex outdoor environments.

We believe our work pioneers a new approach to handling the challenges of data generation through realistic simulations and employing a deep neural network tailored to interpret dense point cloud maps directly. While our method adeptly utilizes spatiotemporal robot-centric point cloud maps, it has limitations. Dynamic objects like pedestrians may become embedded within the map, leading to potential prediction inaccuracies.

Looking ahead, we aim to refine our system to produce point-wise continuous traversability estimates rather than relying on a discretized robot footprint box. This enhancement will help to preserve information integrity and provide even more precise traversability estimates for autonomous robot 
navigation.

\newpage

\textit{Statement:} During the parts of the preparation of this work, the author(s) used ChatGPT and Grammarly in order to improve the language and readability. After using this tool/service, the author(s) reviewed and edited the content as needed and take(s) full responsibility for the content of the publication.

\bibliographystyle{elsarticle-num} 
\bibliography{main}
\end{document}